\documentclass[review,3p,times,11pt]{elsarticle}

\usepackage{enumitem}
\usepackage{lineno}
\usepackage[colorlinks,linkcolor=red,anchorcolor=blue,citecolor=green]{hyperref}
\usepackage{multirow}
\usepackage{rotating}
\usepackage{epstopdf}
\usepackage{graphicx}
\usepackage{subfigure}

\usepackage{times}

\usepackage{makeidx}
\usepackage{multirow}
\usepackage{multicol}
\usepackage[dvipsnames,svgnames,table]{xcolor}
\usepackage{graphicx}
\usepackage{epstopdf}
\usepackage{ulem}
\usepackage{url}

\usepackage{array}
\usepackage{amsmath,amssymb,amsfonts}
\usepackage{algorithmic}
\usepackage{algorithm}
\usepackage{graphicx}
\usepackage{textcomp}

\usepackage{epsfig}
\usepackage{makecell}
\usepackage{longtable,booktabs}
\usepackage{longtable}

\usepackage{lscape}

\journal{Journal}

\begin{document}
\begin{frontmatter}
	
	
		
	\title{Heterogeneous Data Fusion Considering Spatial Correlations using Graph Convolutional Networks and its Application in Air Quality Prediction}


\author[label1]{Zhengjing Ma}
\author[label1]{Gang Mei\corref{cor1}}
\ead{gang.mei@cugb.edu.cn}
\cortext[cor1]{Corresponding authors}
\author[label2]{Salvatore Cuomo}
\author[label2]{Francesco Piccialli\corref{cor1}}
\ead{francesco.piccialli@unina.it}
\address[label1]{School of Engineering and Technology, China University of Geosciences (Beijing), 100083, Beijing, China}
\address[label2]{Department of Mathematics and Applications R. Caccioppoli, University of Naples Federico II, Naples, Italy}

\begin{abstract}
	
Heterogeneous data are commonly adopted as the inputs for some models that predict the future trends of some observations. Existing predictive models typically ignore the inconsistencies and imperfections in heterogeneous data while also failing to consider the (1) spatial correlations among monitoring points or (2) predictions for the entire study area. To address the above problems, this paper proposes a deep learning method for fusing heterogeneous data collected from multiple monitoring points using graph convolutional networks (GCNs) to predict the future trends of some observations and evaluates its effectiveness by applying it in an air quality predictions scenario. The essential idea behind the proposed method is to (1) fuse the collected heterogeneous data based on the locations of the monitoring points with regard to their spatial correlations and (2) perform prediction based on global information rather than local information. In the proposed method, first, we assemble a fusion matrix using the proposed RBF-based fusion approach; second, based on the fused data, we construct spatially and temporally correlated data as inputs for the predictive model; finally, we employ the spatiotemporal graph convolutional network (STGCN) to predict the future trends of some observations. In the application scenario of air quality prediction, it is observed that (1) the fused data derived from the RBF-based fusion approach achieve satisfactory consistency; (2) the performances of the prediction models based on fused data are better than those based on raw data; and (3) the STGCN model achieves the best performance when compared to those of all baseline models. The proposed method is applicable for similar scenarios where continuous heterogeneous data are collected from multiple monitoring points scattered across a study area.

\end{abstract}

\end{frontmatter}


%%\tableofcontents
%%\rule[-10pt]{0.35\textheight}{0.05em}

%%\linenumbers
%%\switchlinenumbers

\newpage
\section*{List of Abbreviations }
\begin{table}[htbp]
	%	\begin{center}
	\begin{tabular}{p{35pt}p{250pt}}
		
		Ada$-$DT & Adaptive Boosted Decision Tree  \\
		CNN  &  Convolutional Neural Networks  \\
		ETR & Extra Trees Regressor  \\
		GBRT  &  Gradient Boosted Decision Tree \\
		GCN  &  Graph Convolutional Networks  \\
		GCNN  &  Gated Convolutional Neural Networks  \\
		GLU & Gated Linear Unit  \\
		GRU & Gated Recurrent Units  \\
		IoT  &  Internet of Things \\
		LSTM & Long Short Term Memory  \\
		MAE & Mean Absolute Error  \\
		MAPE & Mean Absolute Percentage Error  \\
		RBF  &  Radial Basis Functions  \\
		RF  & Random Forest   \\
		RMSE & Root Mean Squared Errorss  \\
		RNN & Recurrent Neural Networks  \\
		STGCN   & Spatio-Temporal Graph Convolutional Network \\
		T$-$GCN & Temporal Graph Convolution Networks  \\
		Xgboost  &  eXtreme Gradient Boosting  \\

	\end{tabular}
	%	\end{center}
\end{table}

\newpage

%\newpage

%\linenumbers

\section{Introduction}
\label{sec1}

In the era of information explosion, with the advancements in the Internet of Things (IoT) \cite{ren2017serving}, a large amount of heterogeneous data is collected from multiple points that belong to different monitoring sources and are scattered in different locations \cite{meng2020survey}. However, the complexities of heterogeneous data result in difficulties with regard to information exchange and decision making. To leverage these existing heterogeneous data and transform them into reliable and accurate information, data fusion is commonly used to address the problem of fusing heterogeneous data.

Data fusion fuses raw data collected from multiple monitoring points to obtain improved information \cite{hall1997introduction,Ngiam2011689}. Typically, the fused information provides a more accurate description than independent observations and increases the robustness and confidence of the data application; the fused data are thus considered to have more complete value than the individual raw observations \cite{barcelo2019self,Gite201685}. Furthermore, data fusion is capable of extending both the temporal and spatial coverage information and further decreasing the ambiguities and uncertainties of multiple monitoring station measurements \cite{Guo2019215}. Data fusion is widely applied in the fields of air traffic control, robotics, manufacturing, medical diagnostics, and environmental monitoring.

A typical application scenario involving heterogeneous data sources is the field of environmental monitoring. In this scenario, a common application for the heterogeneous data collected is air quality prediction. Air quality prediction provides information for government measures, including traffic restrictions, plant closures, and restrictions on outdoor activities. Typically, different data from multiple distinct monitoring stations are required to develop a prediction model \cite{Wong2004404}. Air quality depends on external factors, and one of the most important factors is the set of meteorological conditions that vary over time and space. This implies that having a large amount of heterogeneous data related to meteorological conditions from meteorological monitoring stations is crucial for air quality prediction. In recent years, as state-of-the-art technology, machine learning algorithms have been gradually introduced to develop models for air quality prediction using given datasets. The commonly utilized dataset consists primarily of heterogeneous data collected from diverse air quality monitoring stations and meteorological monitoring stations located in a study area \cite{He201811,Enebish2020,Elangasinghe2014106,Maharani2019}.

Although this kind of predictive model performs well, it suffers from several limitations.

First, for these common predictive models, a typical practice is that meteorological station data (e.g., temperature, humidity, wind speed, wind direction, etc.) from the vicinity of the air quality monitoring stations are directly taken as input. However, the data collected from meteorological stations have diverse spatial and temporal distributions that result from the influence of each external factor and some complicated interactions from combinations of external factors. These common models ignore the effects of the complex multivariate relationships among various monitoring targets to different degrees. These relationships are reflected in spatial and temporal correlations.

Second, these predictive models primarily focus on local information and commonly adopt a limited number of air quality monitoring stations for prediction. In reality, it is difficult to present these local predictions as an indication of future trends for the entire study area.

Third, previous research studies primarily considered their study areas as grid-based regions to acquire geospatial information \cite{Le202055,Zhang20194341}. Although the required computation is relatively fast when partitioning a research area into grids, it is challenging to extract irregular spatial correlations from these regular rectangular areas. Convolutional neural networks (CNNs) are effective in extracting local patterns from spatially correlated data; however, they are generally applicable to standard grid-based data and are not suitable for non-grid data.

To address the above problems, in this paper, considering the spatial correlations among monitoring points (e.g., air quality monitoring stations), we propose a deep learning method for fusing heterogeneous data collected from multiple monitoring points using graph convolutional networks (GCNs) to predict future trends of some observations. The essential idea behind the proposed method is first to fuse the heterogeneous data collected based on the locations of the monitoring points and then to perform prediction based on global information (i.e., fused information from all monitoring points within a study area) rather than local information (i.e., information from a single monitoring point within a study area).

The contributions of this paper can be summarized as follows.

(1)	We consider the spatial distribution of monitoring points and fuse the heterogeneous data collected from these monitoring points that contain spatial information using the RBF-based method.

(2) We connect all monitoring points in the entire study area to construct a fully connected graph and thus replace the traditional single monitoring point or few monitoring points with graph-structured data.

(3) We employ a deep learning model called the STGCN that combines gated convolutional neural networks (GCNNs) and the GCN layer utilizing a particular structure to predict the future trends of some observations [30]. The GCN layer captures spatial correlations between the constructed graph-structured data. The GCNN layer captures temporal correlations between the constructed sequence data.

(4) We evaluate the proposed method by applying it for air quality prediction based on a real-world dataset collected from multiple monitoring points.

The rest of this paper is organized as follows. Section \ref{sec:2:methods} describes the proposed method in detail. Section \ref{sec:3:results} applies the method in a real case and analyzes the results. Section \ref{sec:4:discussion} discusses the advantages and shortcomings of the proposed method, and the potential future work. Section \ref{sec:5:conclusion} concludes the paper.

\section{Methods}
\label{sec:2:methods}
\subsection{Overview}
In this paper, we propose a deep learning method for fusing heterogeneous data collected from multiple monitoring points using GCNs to predict the future trends of several observations. First, we obtain and clean the raw heterogeneous data from multiple monitoring points scattered over the study area. Second, we fuse the cleaned data using the proposed RBF-based method by considering the spatial correlations. Third, we construct the spatially and temporally correlated data based on the above fused data as the input for further prediction. Fourth, based on the constructed spatiotemporal data, we build a novel deep learning model called the spatiotemporal graph convolutional network (STGCN) for prediction. The predictive model combines GCN layers that can be used to capture the spatial correlations between the constructed data and GCNN layers that can be used to capture the temporal correlations between the constructed data.

\begin{figure*}[!ht]
	\centering
	\includegraphics[width=14cm]{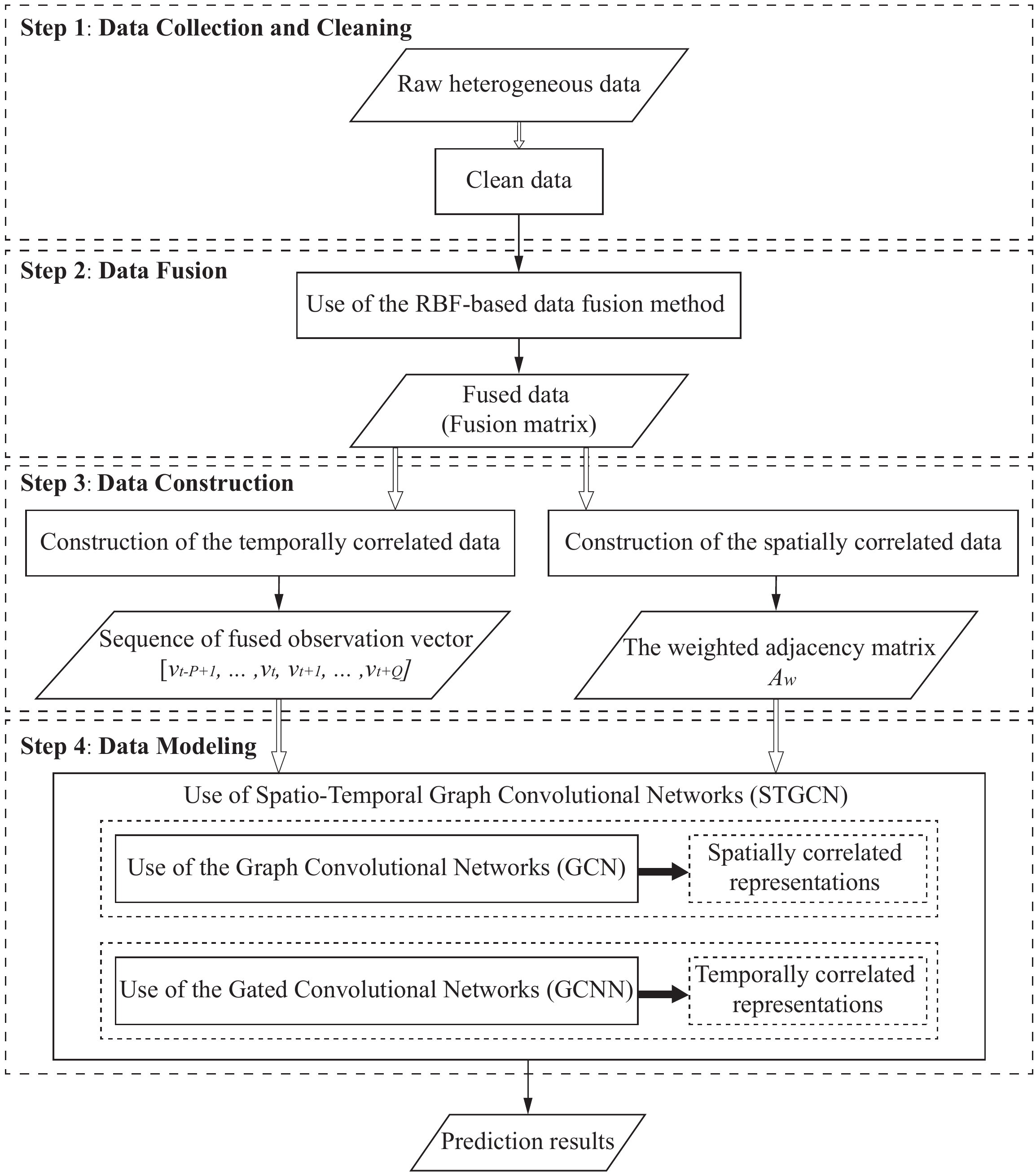}
	\caption{The workflow of the proposed method} 
	\label{Figure1}
\end{figure*}

\subsection{Step 1: Data Collection and Cleaning}
\label{subsec:1:Data Collection and Cleaning}

With the advancement of IoT technology, the number of monitoring sources is increasing rapidly. As a result, the number of monitoring targets is rapidly increasing, and large amounts of heterogeneous data are generated within a short period of time. When utilizing heterogeneous data collected from multiple points, the data should first be cleaned to extract valuable information from these raw data. Specifically, the first step is checking the quality of the data to identify incorrect, inconsistent, missing, or redundant information. The second step is to clean (i.e., denoise) the data, including removing incorrect values and dealing with missing values.

\subsection{Step 2: Data Fusion}
\label{sec:2.4:Datafusions}

We assume that there are a total number of $S$ monitoring points scattered across a study area. For a given types of monitoring source $m$, a study area has $K_{m}$ monitoring targets that vary with the type of monitoring sources. At each time point, for a given monitoring point $s$ that belongs to monitoring source $m$, $C_{k,m}(s)$ denotes one of its observations, and $k$ represents the $k^{th}$ monitoring target. In the entire study area, all monitoring sources monitor a total of $K$ targets, $K=\sum K_{m}$. 

Typically, a pair of monitoring points with a short distance between them have a stronger correlation than that of a pair with a long distance between them \cite{tobler1970computer}. Based on this conception, we propose a method to fuse heterogeneous data collected from these monitoring points. The proposed method considers the spatial location of each monitoring point in a study area. We incorporate the spatial correlations between these independent observations into a distance-based fusion method, thus fusing the data from one set of monitoring points with data from the other set of monitoring points according to spatial information, both of which belong to different monitoring sources, according to spatial information.

In a given study area, we assume that a set of $N_{j}$ monitoring points $s_{j}$, $j=1,2, \ldots, N_{j}$ belongs to a type of monitoring source $m_{j}$, and its $k_{j}^{th}$ observation is $C_{k_{j},m_{j}}$. Furthermore, there is also another set of $N_{i}$ monitoring points $s_{i}$, $i=1,2, \ldots, N_{i}$, belonging to another type of monitoring source $m_{i}$, and its $k_{i}^{th}$ observation is $C_{k_{i},m_{i}}$. Based on the distance $dist(i, j)$ between  monitoring points $s_{i}$ and $s_{j}$, we fuse the observation $C_{k_{i},m_{i}}(s_{i})$ of point $s_{i}$ into point $s_{j}$ and fuse the observations $C_{k_{j},m_{j}}(s_{j})$ of  point $s_{j}$ into point $s_{i}$. Therefore, for the heterogeneous data collected from scattered monitoring points, a function $F$ can be employed to perform fusion. When $F(s_{j}) = C_{k_{j},m_{j}}(s_{j})$, we have $F(s_{i}) = C_{k_{j},m_{j}}(s_{i})$.

\begin{figure*}[!ht]
	\centering
	\includegraphics[width=14cm]{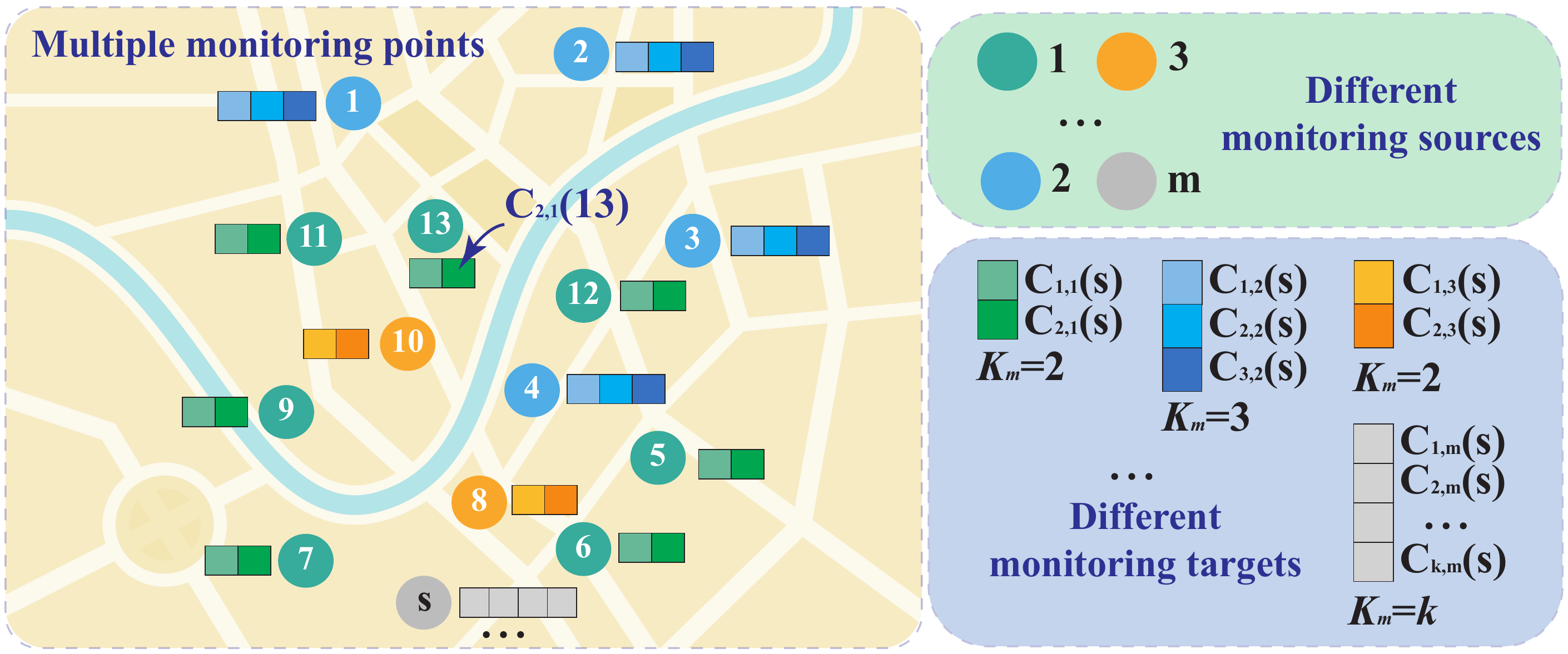}
	\caption{An example of specific scenarios that heterogeneous data was collected from multiple monitoring points scattered in a study area. Here, $C_{2,1}(13)$ represent observation of $2^{nd}$ monitoring targets from given monitoring point $s_{13}$, the type of the monitoring source is 1. Different observation describe using diverse colors.} 
	\label{Figure2}
\end{figure*}

A linear combination of RBFs is able to construct a function $F$ of these monitoring points (see Eq. (\ref{equ:eq1})). As a distance-based function, RBFs are particularly applicable to weight and fuse large amounts of data from the spatial correlations between multiple scattered monitoring points \cite{Boyd20101435}. In this paper, we employ an RBF-based method to perform fusion. Here, we adopt the Gaussian basis function, i.e., the Gaussian RBF, as our RBF (see Eq. (\ref{equ:eq2})).

\begin{equation}
\label{equ:eq1}
F=\sum_{j=1}^{N_{j}} w_{j} \phi_{j}(d i s t)
\end{equation}

\begin{equation}
\label{equ:eq2}
\phi=\exp \left(-c \cdot d i s t^{2}\right)
\end{equation}
where $\phi$ is the Gaussian RBF. $dist$ represents the Euclidean distance. For two monitoring points $s_{i}$ and $s_{j}$, the distance  $dist(i, j)=\left\|s_{i}-s_{j}\right\|$. The weights $w_{j}=\left[w_{1}, w_{2}, \ldots, w_{N_{j}}\right]^{\mathrm{T}}$ can be estimated by solving the linear system of equations.

First, we calculate the Euclidean distances $dist$ between the above $N_{j}$ monitoring points and obtain a distance matrix of size $N_{j} \times N_{j}$. Then, these distances are input into the Gaussian RBF to obtain a coefficient matrix $A$, and the size of $A$ is also $N_{j} \times N_{j}$ (see Eq. (\ref{equ:eq3})). Next, the observations from the $N_{j}$ monitoring points are defined as a vector $B=\left[C_{k j, m j}\left(s_{1}\right), \quad C_{k j, m j}\left(s_{2}\right),\ldots, C_{k j, m j}\left(s_{N j}\right)\right]^{\mathrm{T}}$. Thus, a linear group of equations $A w_{j}=B$ is obtained, thus leading to the aforementioned weights $w_{j}$. Finally, the above $N_{i}$ monitoring points yield $k_{j}$ observations, i.e., fused data, where $k_{j}$ denotes the $k_{j}^{th}$ monitoring target of the monitoring source $m_{j}$ (see Eq. (\ref{equ:eq4})).

\begin{equation}
\label{equ:eq3}
A=\left[\begin{array}{cccc}
\phi\left(\operatorname{dist}\left(s_{1}, s_{1}\right)\right) & \phi\left(\operatorname{dist}\left(s_{1}, s_{2}\right)\right) & \cdots & \phi\left(\operatorname{dist}\left(s_{1}, s_{N_{j}}\right)\right) \\
\phi\left(\operatorname{dist}\left(s_{2}, s_{1}\right)\right) & \phi\left(\operatorname{dist}\left(s_{2}, s_{2}\right)\right) & \cdots & \phi\left(\operatorname{dist}\left(s_{2}, s_{N_{j}}\right)\right) \\
\vdots & \vdots & \ddots & \vdots \\
\phi\left(\operatorname{dist}\left(s_{N_{j}}, s_{1}\right)\right) & \phi\left(\operatorname{dist}\left(s_{N_{j}}, s_{2}\right)\right) & \cdots & \phi\left(\operatorname{dist}\left(s_{N_{j}}, s_{N_{j}}\right)\right)
\end{array}\right]
\end{equation}

\begin{equation}
\label{equ:eq4}
C_{k j, m_{j}}\left(s_{i}\right)=\sum_{j=1}^{N_{j}} w_{j} \phi_{j}(\operatorname{dist}(i, j)), \quad i=1,2, \cdots, N_{i}
\end{equation}

After performing the RBF-based fusion process, a fusion matrix (FM) is obtained for a single time step (see Eq. (\ref{equ:eq5})). Each row of the matrix represents a monitoring point, and each column represents a monitoring target. There are $S$ monitoring points and $K$ monitoring targets in a study area. For multiple time steps $T$, the size of the fusion matrix is $T \times S \times K$. The fusion matrix can be used to construct spatiotemporally correlated data.

\begin{equation}
\label{equ:eq5}
F M=\left(\begin{array}{ccc}
C_{1}(1) & \cdots & C_{K}(1) \\
\vdots & \ddots & \vdots \\
C_{1}(S) & \cdots & C_{K}(S)
\end{array}\right)
\end{equation}

\begin{figure*}[!ht]
	\centering
	\includegraphics[width=14cm]{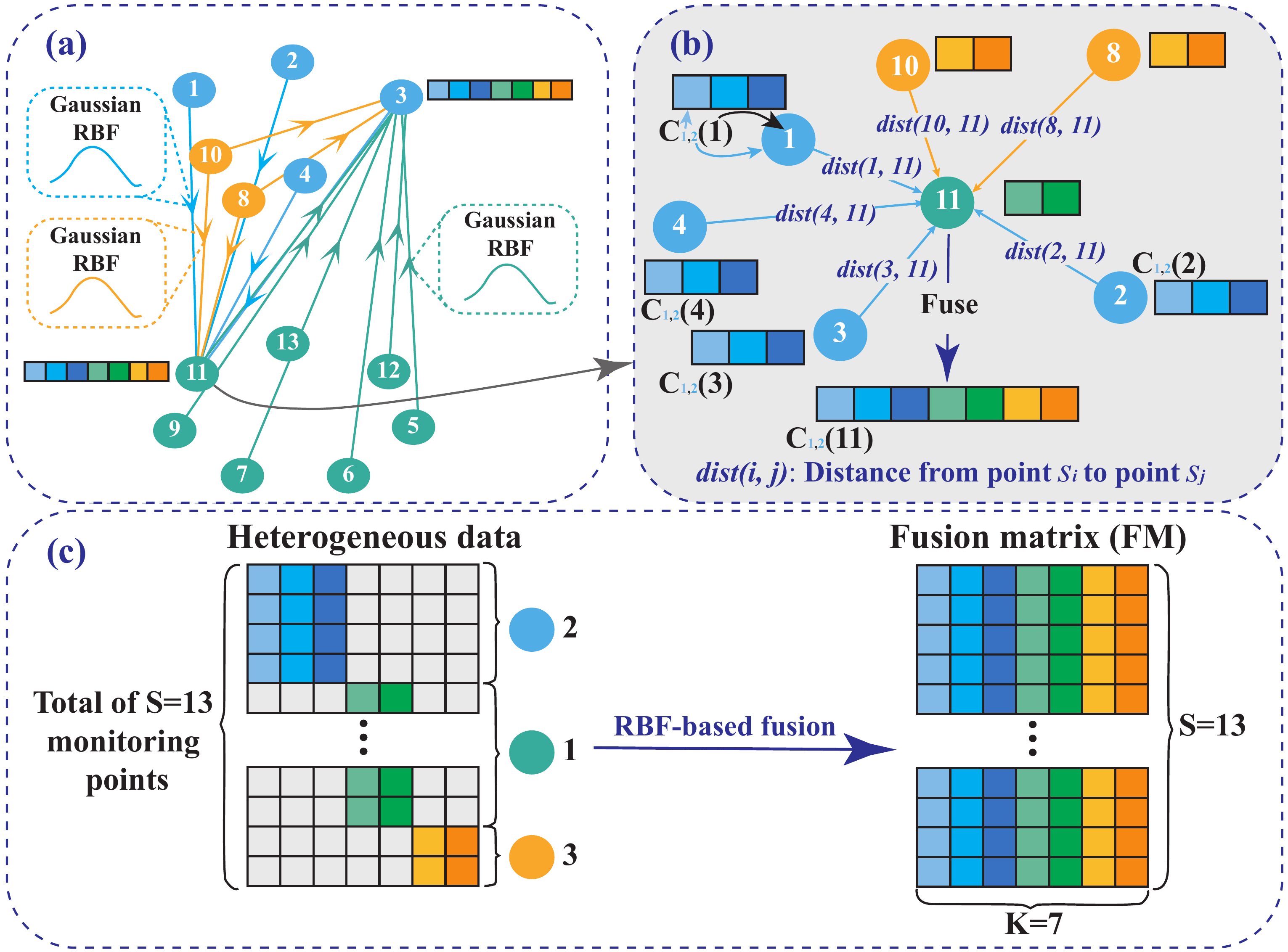}
	\caption{An illustration of the process of the proposed RBF-based fusion. (a) An example of performing the RBF-based fusion in monitoring point $s_{3}$ and monitoring point $s_{11}$. These monitoring points obtain fused data from surrounding neighbors. (b) A more detailed illustration of the process of obtaining fused data for monitoring point  $s_{11}$, i.e., the fusing process is based on the distance between the point and its neighbors. (c) Establishment of the fusion matrix with size of $S \times K$. } 
	\label{Figure3}
\end{figure*}

The detailed process of RBF-based fusion is illustrated in Figure \ref{Figure2} and Figure \ref{Figure3}. We assume that there are 13 monitoring points distributed in the given study area. These monitoring points belong to three different monitoring sources; different sources are represented by different colors: green for monitoring source 1, blue for monitoring source 2, and yellow for monitoring source 3. The number of monitoring targets depends on the type of monitoring source. Here, monitoring source 1 has two monitoring targets, monitoring source 2 has three monitoring targets, and monitoring source 3 has two monitoring targets, i.e., there are seven monitoring targets total in the study area.

We take monitoring point $s_{11}$ as an example to introduce how to perform the RBF-based fusion. As illustrated in Figure \ref{Figure2}, $s_{11}$ belongs to monitoring source $1$, and thus, the purpose of the fusion operation is to fuse the observations of both monitoring source $2$ (containing three monitoring targets) and monitoring source $3$ (containing two monitoring targets) into the data of $s_{11}$. 

First, we calculate the Euclidean distances between the $4$ monitoring points that belong to monitoring source $2$ and thus obtain a distance matrix and a coefficient matrix $A_{2}$ with sizes of $4 \times 4$. The coefficient matrix $A_{2}$ is:

\begin{equation}
\label{equ:eq6}
A_{2}=\left[\begin{array}{llll}
\phi(\operatorname{dist}(1,1)) & \phi(\operatorname{dist}(1,2)) & \phi(\operatorname{dist}(1,3)) & \phi(\operatorname{dist}(1,4)) \\
\phi(\operatorname{dist}(2,1)) & \phi(\operatorname{dist}(2,2)) & \phi(\operatorname{dist}(2,3)) & \phi(\operatorname{dist}(2,4)) \\
\phi(\operatorname{dist}(3,1)) & \phi(\operatorname{dist}(3,2)) & \phi(\operatorname{dist}(3,3)) & \phi(\operatorname{dist}(3,4)) \\
\phi(\operatorname{dist}(4,1)) & \phi(\operatorname{dist}(4,2)) & \phi(\operatorname{dist}(4,3)) & \phi(\operatorname{dist}(4,4))
\end{array}\right]
\end{equation}

Second, $B_{2}=\left[C_{1,2}(1), C_{1,2}(2), C_{1,2}(3), C_{1,2}(4)\right]^{\mathrm{T}}$ are observations from the $1^{st}$ monitoring target of monitoring source 2. We can solve for $w$ from the linear equation $A_{2} w=B_{2}$:

\begin{equation}
\label{equ:eq7}
\underbrace{\left[\begin{array}{cccc}
	\phi\left(\operatorname{dist}\left(1, 1\right)\right)& \phi\left(\operatorname{dist}\left(1, 2\right)\right)& \phi\left(\operatorname{dist}\left(1, 3\right)\right)&  \phi\left(\operatorname{dist}\left(1, 4\right)\right)\\
	\phi\left(\operatorname{dist}\left(2, 1\right)\right)& \phi\left(\operatorname{dist}\left(2, 2\right)\right)& \phi\left(\operatorname{dist}\left(2, 3\right)\right)& \phi\left(\operatorname{dist}\left(2, 4\right)\right)\\
	\phi\left(\operatorname{dist}\left(3, 1\right)\right)& \phi\left(\operatorname{dist}\left(3, 2\right)\right)&  \phi\left(\operatorname{dist}\left(3, 3\right)\right)&  \phi\left(\operatorname{dist}\left(3, 4\right)\right)\\
	\phi\left(\operatorname{dist}\left(4, 1\right)\right)&  \phi\left(\operatorname{dist}\left(4, 2\right)\right)&  \phi\left(\operatorname{dist}\left(4, 3\right)\right)& \phi\left(\operatorname{dist}\left(4, 4\right)\right)
	\end{array}\right]}_{A_{2}} 
	\underbrace{\left[\begin{array}{c}
	w_{1} \\
	w_{2} \\
	w_{3} \\
	w_{4}
	\end{array}\right]}_{w}=
\underbrace{\left[\begin{array}{c}
	C_{1,2}(1) \\
	C_{1,2}(2)  \\
	C_{1,2}(3)  \\
	C_{1,2}(4) 
	\end{array}\right]}_{B_{2}}
\end{equation}

Third, the distances of these four points from monitoring source 2 to monitoring point $s_{11}$ are calculated as $dist(1,11)$, $dist(2,11)$, $dist(3,11)$, and $dist(4,11)$ (see Figure \ref{Figure3}). 

Finally, by inputting these distances into the Gaussian RBF (see Eq. (\ref{equ:eq4})), the result $C_{1,2}(11)$ is obtained, in which monitoring point $s_{11}$ fuses the data from the $1^{st}$ monitoring target of monitoring source 2. As illustrated in Figure \ref{Figure3}, by repeating this process, a fusion matrix with a size of $13 \times 7$ is obtained at a single time step. Here, 13 represents 13 monitoring points and 7 represents 7 monitoring targets. After adding the temporal information, the assembled fusion matrix can be used to construct spatiotemporal correlation data for further model predictions.

\subsection{Step 3: Data Construction}
\label{sec:2.5:Dataconstruction}
In this section, we introduce the process of data construction for spatially and temporally correlated data.
\subsubsection{Construction of Spatially Correlated Data}

We adopt a weighted fully connected graph $G=\left(V_{t}, E, A_{w}\right)$ to represent the non-Euclidean spatial correlations among multiple monitoring points, where $V_{t}$ is a set of nodes corresponding to the data collected from $S$ monitoring points in a study area. Each node is connected to every other node, implying the existence of correlations between all monitoring points in the entire study area (each pair of nodes has an edge $E$). The number of edges is $S(S-1) / 2$. A weighted adjacency matrix $A_{w}$ is used to represent the similarities between nodes: $A_{w} \in \mathbb{R}^{S^{\times} S}$. $A_{w}$ is defined as in Eq. (\ref{equ:eq9}). In the matrix, each element is the weight $w_{ij}$ of each edge $e_{ij}$, representing the spatial correlation between two nodes (i.e., monitoring points $s_{i}$ and $s_{j}$). The weight is calculated using Gaussian similarity functions based on the distance between both nodes. A greater weight implies a stronger correlation between the two nodes.

\begin{equation}
\label{equ:eq8}
\left.w(i, j)=\exp \left(-dist(i, j)^{2} / \sigma^{2}\right)\right)
\end{equation}

\begin{equation}
\label{equ:eq9}
A_{w}=\left(\begin{array}{cccc}
0 & w(1,2) & \cdots & w(1, S) \\
w(2,1) & 0 & \cdots & w(2, S) \\
\vdots & \vdots & \ddots & \vdots \\
w(S, 1) & w(S, 2) & \cdots & 0
\end{array}\right)
\end{equation}
where $dist(i ,j)$ denotes the distance between monitoring points $s_{i}$ and $s_{j}$, and $\sigma$ is the standard deviation of distances, which controls the width of the neighborhoods \cite{VonLuxburg2007395}. 

\begin{figure*}[!ht]
	\centering
	\includegraphics[width=14cm]{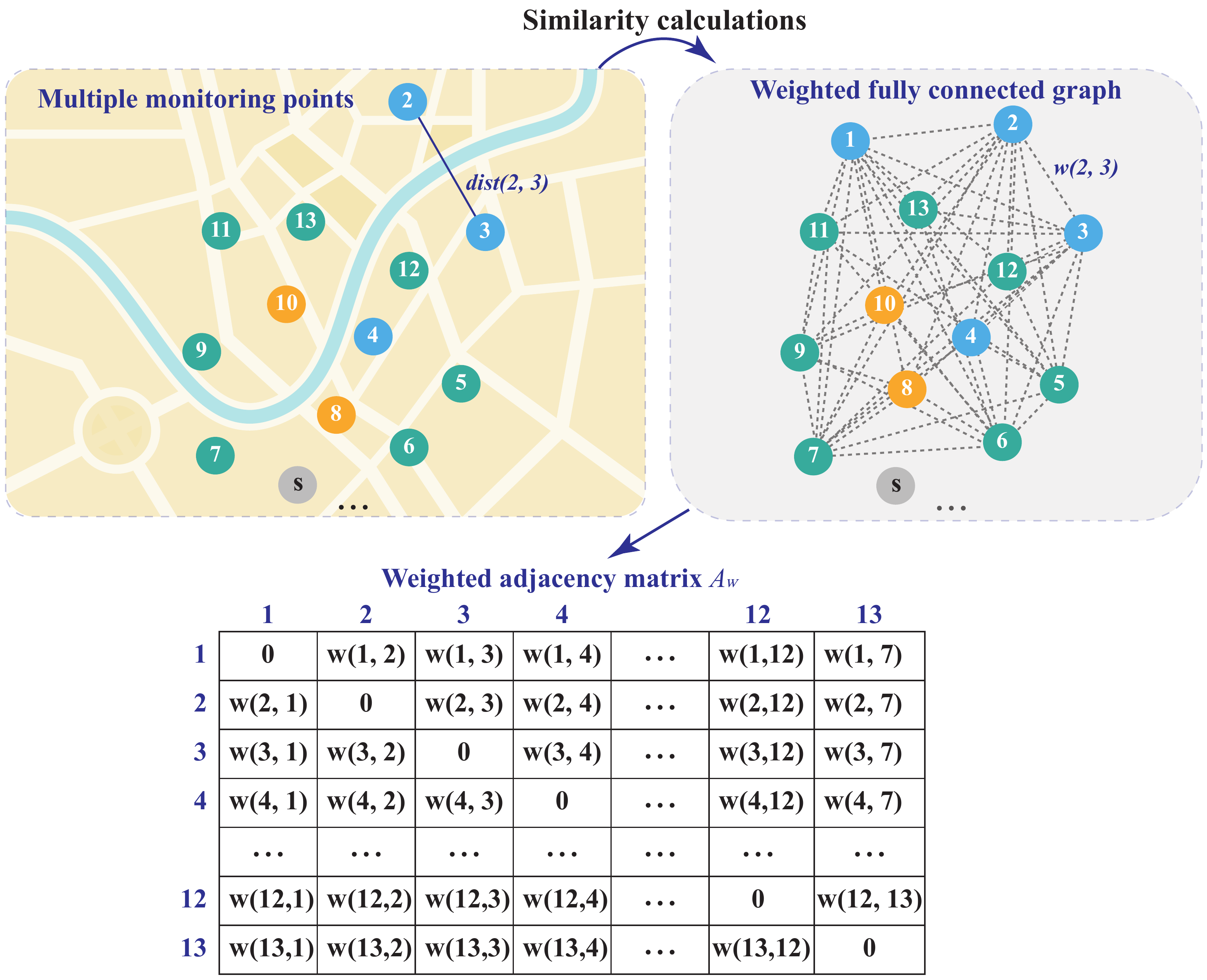}
	\caption{An example illustrating the process of constructing a weighted fully connected graph to represent all monitoring points in the entire study area. There are 13 monitoring points. We construct a fully connected graph to represent these points and then calculate the similarity between the two connected nodes (i.e., monitoring points) in the graph using Gaussian similarity to obtain a weighted adjacency matrix.} 
	\label{Figure4}
\end{figure*}

\subsubsection{Construction of Temporally Correlated Data}

To represent the temporal correlations among heterogeneous data collected from multiple monitoring points, we construct a fused observation vector $v_{t} \in \mathbb{R}^{S}$ of $S$ monitoring points at time step $t$, where each element records historical observations (i.e., fused data from monitoring targets) for a monitoring point. Specifically, a frame $v_{t} \in \mathbb{R}^{S^{\times} K}$ denotes the current status of the fusion matrix at time step $t$. The corresponding information is stored in the fully connected graph $G$. A typical time series prediction problem requires the use of data from the previous $P$ time steps to predict data for the next $Q$ time steps. Based on vector $v_{t}$, the prediction problem can be defined as in Eq. (\ref{equ:eq10}).

\begin{equation}
\label{equ:eq10}
\hat{v}_{t+1}, \ldots, \hat{v}_{t+Q}=\underset{v_{t+1}, \cdots, v_{t+Q}}{\arg \max } \log P\left(v_{t+1}, \ldots, v_{t+Q} \mid v_{t-P+1}, \ldots, v_{t}\right)
\end{equation}

There are $S$ monitoring points in a study area that monitor a total of $K$ targets. Then, $\left[v_{t-P+1}, \cdots, v_{t}\right] \in \mathbb{R}^{P \times S \times K}$  is fusion matrix FM for the previous $P$ time steps. $\left[\hat{v}_{t+1}, \cdots, \hat{v}_{t+Q}\right] \in \mathbb{R}^{P \times S \times K^{pre}}$ contains the prediction results of $K^{pre}$ monitoring targets for the next $Q$ time steps.

\begin{figure*}[!ht]
	\centering
	\includegraphics[width=14cm]{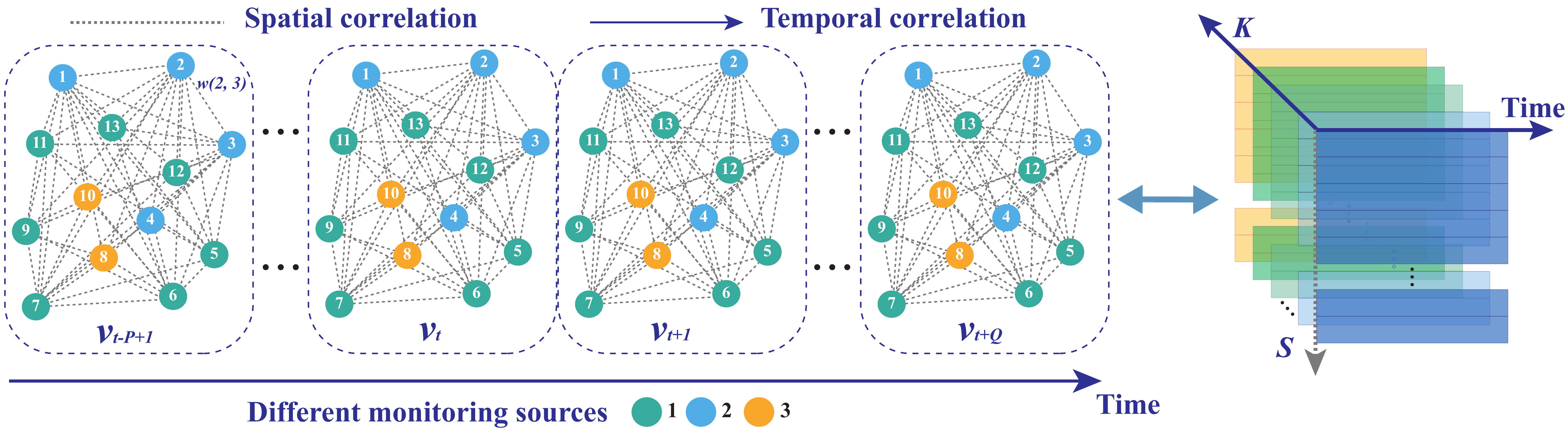}
	\caption{An example illustrating temporally correlated data. At time step $t$, each current status of all monitoring targets from all monitoring points in the entire study area is recorded in graph-structured data.} 
	\label{Figure5}
\end{figure*}

\subsection{Step 4: Data Modeling}

In this section, we introduce details of a novel deep learning model called the spatiotemporal graph convolutional network (STGCN), which adopts the constructed spatiotemporal data as input. First, we introduce the GCN that is used to capture spatially correlated representations from the constructed graph-structured data. Second, we introduce the GCNN that can be used to capture temporally correlated representations from the constructed sequence data. Finally, we introduce the STGCN, which fuses the spatial and temporal representations to predict the future trends of some observations.

\subsubsection{Use of the Graph Convolutional Network (GCN)}

In this section, we introduce how to use GCN to capture spatially correlated representations from the constructed graph-structured data.

Recently, GCNs have attracted increasing attention; they generalize conventional convolution for data with non-Euclidean structures \cite{Hammond2011129} and are thus perfectly suitable for extracting spatial correlations from graph-structured data. Structurally, through action on the nodes of the fully connected graph $G$ and their neighbors, the GCN is able to capture spatial correlations between the nodes and their surrounding nodes and encode the attributes of the nodes (i.e., different information stored at each node).

There are two common types of methods for performing graph convolution, i.e., spatial and spectral methods; in this paper, we focus on the latter type. Specifically, graph convolution operations learn the spatially correlated representations of graph-structured data in the spectral domain using graph Fourier transforms. According to the theory of spectral graph convolution, the graph convolution operator can be defined as $*\mathcal{G}$, which denotes the multiplication between graph signals $x \in \mathbb{R}^{S}$  with a kernel $\Theta$. A Fourier basis  $U \in \mathbb{R}^{S \times S}$ acts on the nodes of the fully connected graph $G$ and their first-order neighbors to capture the spatial correlations among these nodes. Consequently, the convolutional process of a GCN can be described as follows: a graph signal $x$ is filtered by a kernel $\Theta$  with multiplication between the kernel $\Theta$ and the graph Fourier transform $U^{T} x$ (see Eqs. (\ref{equ:eq11}) and (\ref{equ:eq12})) \cite{Shuman201383}. The structure of the GCN is illustrated in Figure \ref{Figure6}.

\begin{equation}
\label{equ:eq11}
\Theta *\mathcal{G} x=\Theta(L) x=\Theta\left(U \Lambda U^{T}\right) x=U \Theta(\Lambda) U^{T} x
\end{equation}

\begin{equation}
\label{equ:eq12}
L=I_{n}-D^{-\frac{1}{2}} A_{w} D^{-\frac{1}{2}}=U \Lambda U^{T} \in \mathbb{R}^{S \times S}
\end{equation}
where $L$ denotes a normalized Laplace matrix. $U \in \mathbb{R}^{S \times S}$ is matrix composed of eigenvectors from $L$. $I_{n}$ denotes an identity matrix. $D \in \mathbb{R}^{S \times S}$ denotes the diagonal degree matrix with $D_{i i}=\Sigma_{j} A_{w}$. Both $\Lambda \in \mathbb{R}^{S \times S}$ and filter $\Theta(\Lambda)$ denotes the diagonal matrix of eigenvalues of $L$. 

To model higher-order neighborhood interactions in the graph, the multiple graph convolutional layers can be stacked, and thus, a deep architecture can be constructed to capture deep spatial correlations \cite{Duvenaud20152224}. The multilayer model requires scaling and normalization. Then, the graph convolution is further expressed as in Eq. (\ref{equ:eq13}).

\begin{equation}
\label{equ:eq13}
\begin{aligned}
\Theta *\mathcal{G} x &=\theta\left(I_{n}+D^{-\frac{1}{2}} A_{w} D^{\frac{1}{2}}\right) x \\
&=\theta\left(\tilde{D}^{-\frac{1}{2}} \tilde{A}_{w} \tilde{D}^{-\frac{1}{2}}\right) x
\end{aligned}
\end{equation}
where $\theta$ is a single shared parameter of the kernel $\Theta$, $A_{w}$ and $D$ are normalized to $\tilde{A}_{w}$, $\tilde{A}$ and $\tilde{D}$; $\tilde{A}_{w}=A_{w}+I$ is a matrix with a self-connection structure, $I$ is the identity matrix. $\tilde{D}$ denotes the diagonal degree matrix with $\tilde{D}_{i i}=\Sigma_{j} \tilde{A}_{w}$.

The above graph convolution operator $*\mathcal{G}$ is mainly applicable to graph signals $x \in \mathbb{R}^{S}$. For a graph signal $X \in \mathbb{R}^{S \times K}$ with $K$ channels ($K$ here refers to the $K$ monitoring targets), $*\mathcal{G}$ is able to be extended to multidimensional tensors, and is defined as in Eq. (\ref{equ:eq14}). 

\begin{equation}
\label{equ:eq14}
\Theta * \mathcal{G} x_{m}=\sum_{k=1}^{K} \Theta_{k, m}(L) x_{m} \in \mathbb{R}^{S}, 1 \leq m \leq C_{o}
\end{equation}
where $L$ denotes a normalized Laplace matrix. $K$ are the input dimensions, $C_{o}$ are the output dimensions. $\Theta \in \mathbb{R}^{u \times K \times C_{o}}$, where $u$ is the kernel size for the graph convolution.

\subsubsection{Use of the Gated Convolutional Neural Networks (GCNN)}

In this section, we introduce how to use GCNNs to capture temporally correlated representations from the constructed sequence data.

Compared to the traditional CNN, a GCNN adds a special gating mechanism that allows it to be used to capture the temporal correlations from time-series data. Compared to the RNN, which is a traditional time series analysis model based on a complex gating mechanism, the GCNN has a simpler structure, which enables it to respond faster for dynamic changes and consequently train faster and independently of previous steps \cite{21}. Furthermore, the GCNN is capable of obtaining the size of the space between each cell (i.e., time step) based on filters, thus allowing it to further capture the relationships between the different observations in the time series data.

\begin{figure*}[!ht]
	\centering
	\includegraphics[width=14cm]{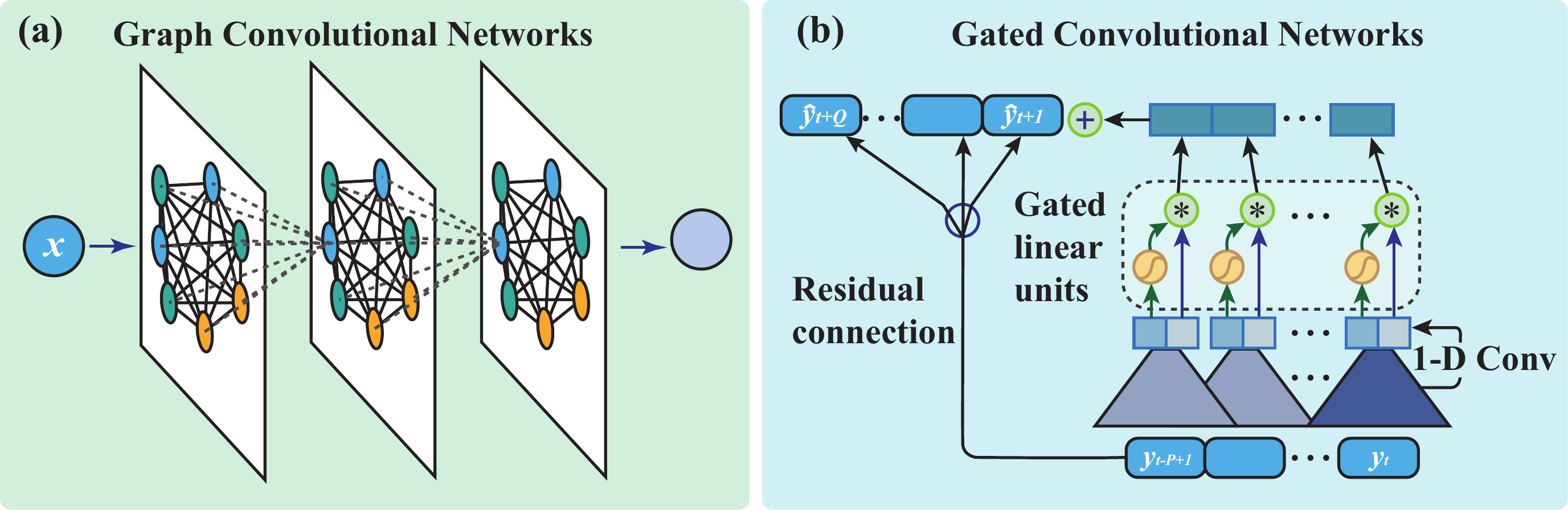}
	\caption{Structures of GCN and GCNN models. (a) The architecture of the Graph Convolution Networks (GCN). (b) Gated Convolutional Networks (GCNN), 1D-Conv represents 1D-convolution.} 
	\label{Figure6}
\end{figure*}

The structure of the GCNN is illustrated in Figure \ref{Figure6}. The GCNN employs a causal convolution as a temporal convolution layer, followed by a gated linear unit (GLU) as a nonlinear activation function. These temporal convolutional layers are stacked using residual connections. The GLU determines which information is passed to the next level. By handling sequence data with a non-recursive approach, the temporal convolution layer is easily computed in parallel and ameliorates the gradient explosion problem that exists in traditional RNN-based learning methods. As a 1D convolution layer (1D-Conv), the temporal convolution layer is convoluted along the temporal dimension with a filter size of $f_{t}$ for exploring $f_{t}$ neighbors of the input elements.

For the constructed sequence data, the input of the temporal convolution layer at each node is considered as a sequence of length $P$. Each node has $K$ channels, and then the input is $y \in \mathbb{R}^{P \times K}$. Therefore, the convolution involves a linear transformation that takes $K$  channels, $f_{t}$ consecutive data at a time, and turns them into output channels. As a result, the length of the output sequence is $f_{t-1}$ less than the length of the input. Given the filter (i.e., the convolution kernel) $\Gamma \in \mathbb{R}^{f_{t} \times K \times 2 C_{o}}$, the temporal convolution can be expressed as Eq. (\ref{equ:eq15}).

\begin{equation}
\label{equ:eq15}
\Gamma * \tau y=\left(y * \Theta_{1}+b_{1}\right) \odot \sigma\left(y * \Theta_{2}+b_{2}\right) \in \mathbb{R}^{\left(P-f_{t}+1\right) \times C_{o}}
\end{equation}
where both $\Theta_{1}$, $\Theta_{2}$, $b_{1}$, and $b_{2}$ are model parameters, $\Theta$ and $b$ denotes kernels and biases, respectively. $*$ denotes convolution operation. $\odot$ is element-wise product. $\sigma$ denotes sigmoid gate, it determines which information of input are relevant to the structure and dynamic evolution for time series data. The same convolution kernel $\Gamma$ is employed for all $S$ nodes in the graph $G$, and thus defined as $\Gamma * \tau Y$. $Y \in \mathbb{R}^{P \times S \times K}$ denotes a fusion matrix of $P$ time steps.

\subsubsection{Use of the Spatio-Temporal Graph Convolutional Networks (STGCN)}
In this section, we introduce how to use the STGCN to predict the future trends of some observations based on the spatial and temporal representations described above. The STGCN is a deep learning model that includes multiple spatio-temporal convolutional blocks (ST-Conv blocks). The STGCN is designed to handle graph-structured time series data for fusing spatial and temporal representations and performing prediction \cite{22}. 

\begin{figure*}[!ht]
	\centering
	\includegraphics[width=14cm]{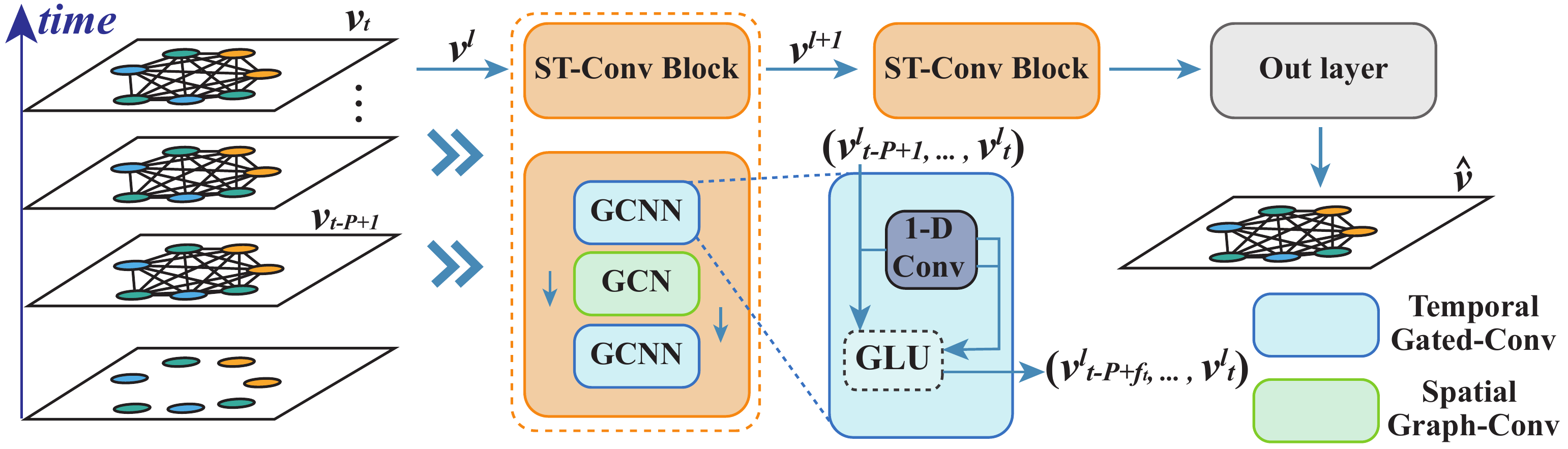}
	\caption{The modeling process of spatiotemporal prediction employing the Spatio-Temporal Graph Convolutional Networks (STGCN)} 
	\label{Figure7}
\end{figure*}

Figure \ref{Figure7} illustrates the specific structure of the ST-Conv block that contains two temporally gated convolution layers and a spatial graph convolution layer is placed between them as a connection. The design decreases the number of channels $K$, thus decreasing the number of parameters involved in the computation and speeding up the training process. Taking sequence data $\left[v_{t-P+1}, \cdots, v_{t}\right] \in \mathbb{R}^{P \times S \times K}$ as a set of inputs, $v^{l} \in \mathbb{R}^{P \times S \times K^{l}}$ is the input of the lth ST-Conv block. $v^{l+1} \in \mathbb{R}^{\left(P-2\left(f_{t}-1\right)\right) \times S \times K^{l+1}}$ is the output, and it can be obtained by Eq. (\ref{equ:eq16}).

\begin{equation}
\label{equ:eq16}
v^{l+1}=\Gamma_{1}^{l} * \tau \operatorname{ReLU}\left(\Theta^{l} *\mathcal{G}\left(\Gamma_{0}^{l} * \tau v^{l}\right)\right)
\end{equation}
where $\Gamma_{0}^{l}$ and $\Gamma_{1}^{l}$ denotes the two temporal layers of $l^{th}$ ST-Conv block from top to bottom, respectively.   $\Theta^{l}$ denotes the kernel for graph convolution. $\operatorname{ReLU}(\cdot)$ is the rectified linear units function.

A typical STGCN stacks two ST-Conv blocks and ends with an extra time convolution and a fully connected layer. Here, the time convolution layer maps the output of the last ST-Conv block to the fully connected layer, which is regarded as an output layer. The prediction results for all monitoring points in the entire study area are returned from this fully connected layer. Moreover, the STGCN measures the performance of the model according to its L2 losses, and the predictive loss function is defined as follows.

\begin{equation}
\label{equ:eq17}
L\left(\hat{v} ; W_{\theta}\right)=\sum_{t}\left\|\hat{v}\left(v_{t-P+1}, \ldots, v_{t}, W_{\theta}\right)-v_{t+1}\right\|^{2}
\end{equation}
where $\hat{v}(\cdot)$ is prediction of model, $v_{t+1}$ denotes ground truth. $P$ is the previous $P$ time steps. $W_{\theta}$ denotes trainable parameters. 

\begin{figure*}[!ht]
	\centering
	\includegraphics[width=14cm]{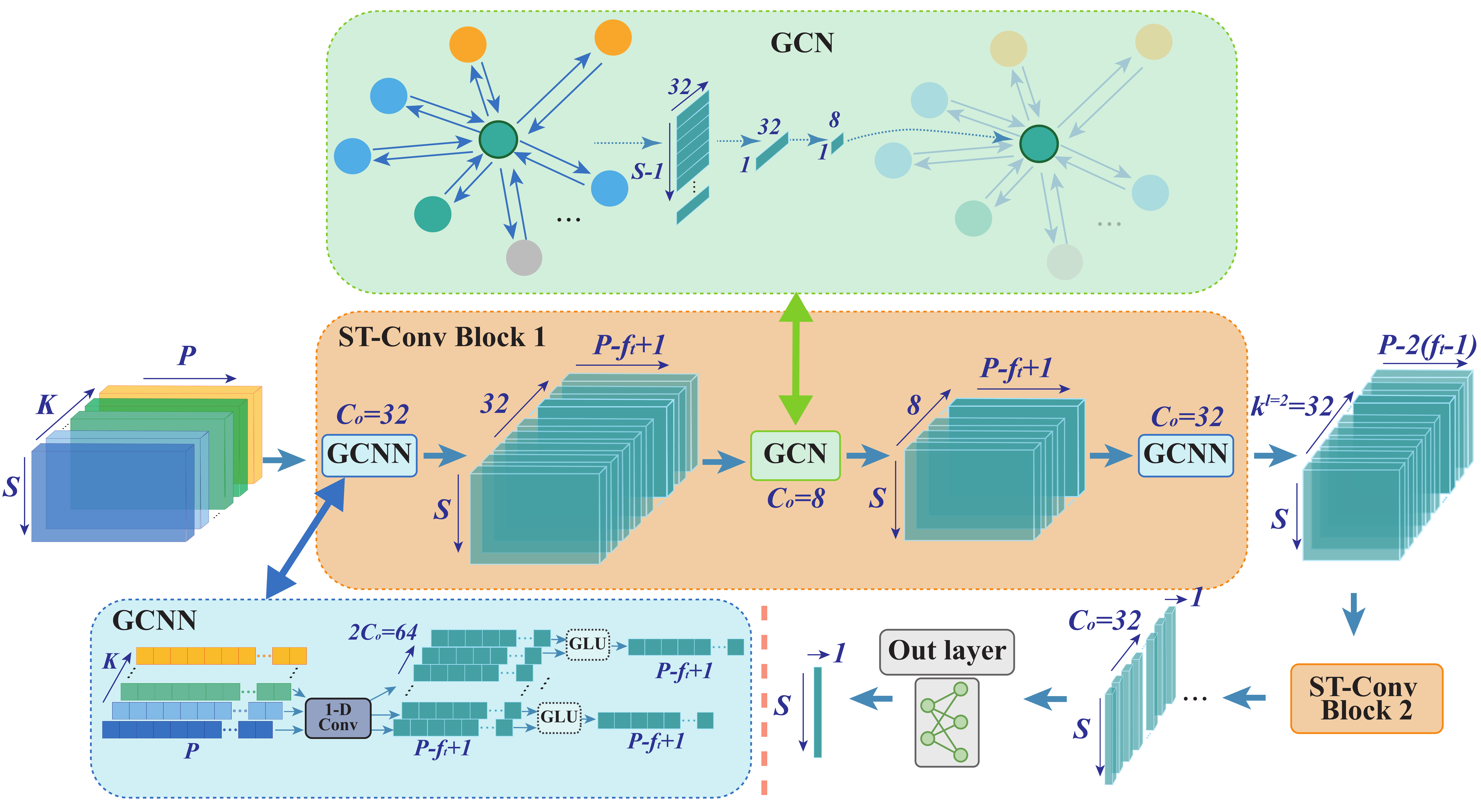}
	\caption{The detailed process of modeling employing Spatio-Temporal Graph Convolutional Networks (STGCN)} 
	\label{Figure8}
\end{figure*}

The detailed modeling process done by employing the STGCN is illustrated in Figure \ref{Figure8}. In this paper, the channels of the three layers in the ST-Conv block are set as 32, 8, and 32. 

First, the input data are fed into the GCNN layer, and the 1-D Conv handles the temporal information of the monitoring points. For each monitoring point with input length $P$, its $K$ input channels are simultaneously convoluted along the time dimension. Consequently, the output length of each point is $P-f_{t}+1$. The convolution kernel $\Gamma \in \mathbb{R}^{f_{t} \times K \times 2 C_{o}}$ in the GCNN maps each input to its individual output. Then, the GLU is used to activate it. After consuming the data with channel 2 and length $P-f_{t}+1$, the GLU generates a single data with length $P-f_{t}+1$. Once all the data have been processed, each node (i.e., monitoring point) in the graph-structured data has 32 channels. 

Next, the input tensor resulting from the temporal convolution is fed into the GCN layer. At each time step, its input for the GCN layer consists of $S$ monitoring points and 32 channels. All monitoring points are connected via a fully connected graph. For a single monitoring point, the remaining $S-1$ monitoring points connected to the point are selected as a subset. The graph convolution operator $*\mathcal{G}$ acts on a subset of monitoring points by reweighting the relevant data. The selected points and the weighting calculation are determined based on the graph-structured data. The input channel for each time step is 32. The number of output channels in the GCN layer is 8. The convolution operation within ST-Conv block 2 is the same as the above process. The spatiotemporal correlations between the constructed data are captured through continuous convolution.

Finally, an output layer (i.e., a fully connected layer) performs the ultimate prediction and then outputs a tensor with a size of $S \times 1$, i.e., the predicted values of $S$ monitoring points at a single time step.

\section{Results: A Real Case}
\label{sec:3:results}
In this section, we apply the proposed method to predict air quality based on a real-world case involving heterogeneous data collected from multiple monitoring points. Moreover, we evaluate and analyze the results.

\subsection{Data Fusion}
In this section, we introduce heterogeneous data fusion based on a real dataset. First, we describe the real dataset in detail and perform RBF-based fusion. Second, we analyze the consistency of the data fusion output using three diverse indicators. Finally, we evaluate the effectiveness of the data fusion process using five machine learning algorithms.

\subsubsection{Data Description}

We collected a real dataset related to air quality in Beijing for the period of January 1, 2017, to February 1, 2018, from the Harvard Dataverse \cite{23}. The dataset consists of two types of monitoring sources: 35 meteorological monitoring stations that collect data related to meteorological conditions (e.g., temperature, humidity, wind speed, etc.) and 17 air quality monitoring stations that collect data related to air quality (e.g., concentrations of various pollutants). These monitoring stations are scattered in different regions throughout the city. Each station records a variety of data every hour. Therefore, the data collected from the multiple monitoring stations in this real case are typical heterogeneous data.

We utilized three observations (i.e., temperature, humidity, and wind speed) from the meteorological monitoring stations, and one observation (i.e., the concentration of PM2.5) from the air quality monitoring stations. The concentration of PM2.5 was adopted as the predicted value. Figure \ref{fig:subfig:13a} illustrates the layout of a total of 53 monitoring stations. The frequency of the measurements was once an hour. The dataset was also aggregated into hourly intervals. Each observation for each station contains 8784 records.

We employed the proposed RBF-based data fusion method at each time step t. After cleaning the data, we utilized linear interpolation to impute missing values for three consecutive hours. Subsequently, we fused the data from the 35 air quality monitoring stations and 17 meteorological monitoring stations with a total of four monitoring targets into a fusion matrix of size $53 \times 4$. 53 denotes the total number of monitoring stations, and 4 denotes the total number of monitoring targets.

\subsubsection{Consistency Analysis of Data Fusion}

We analyzed the consistency of data fusion using three indicators. First, we compared the variances of the raw and fused data. Second, we compared the distributions of the kernel functions for the raw and fused data. Third, we compared the raw and fused data in terms of their time series distributions for different monitoring targets.

\begin{figure}[!ht]
	\centering
	\subfigure[]{
		\label{fig:subfig:9a} 
		\includegraphics[scale=0.5]{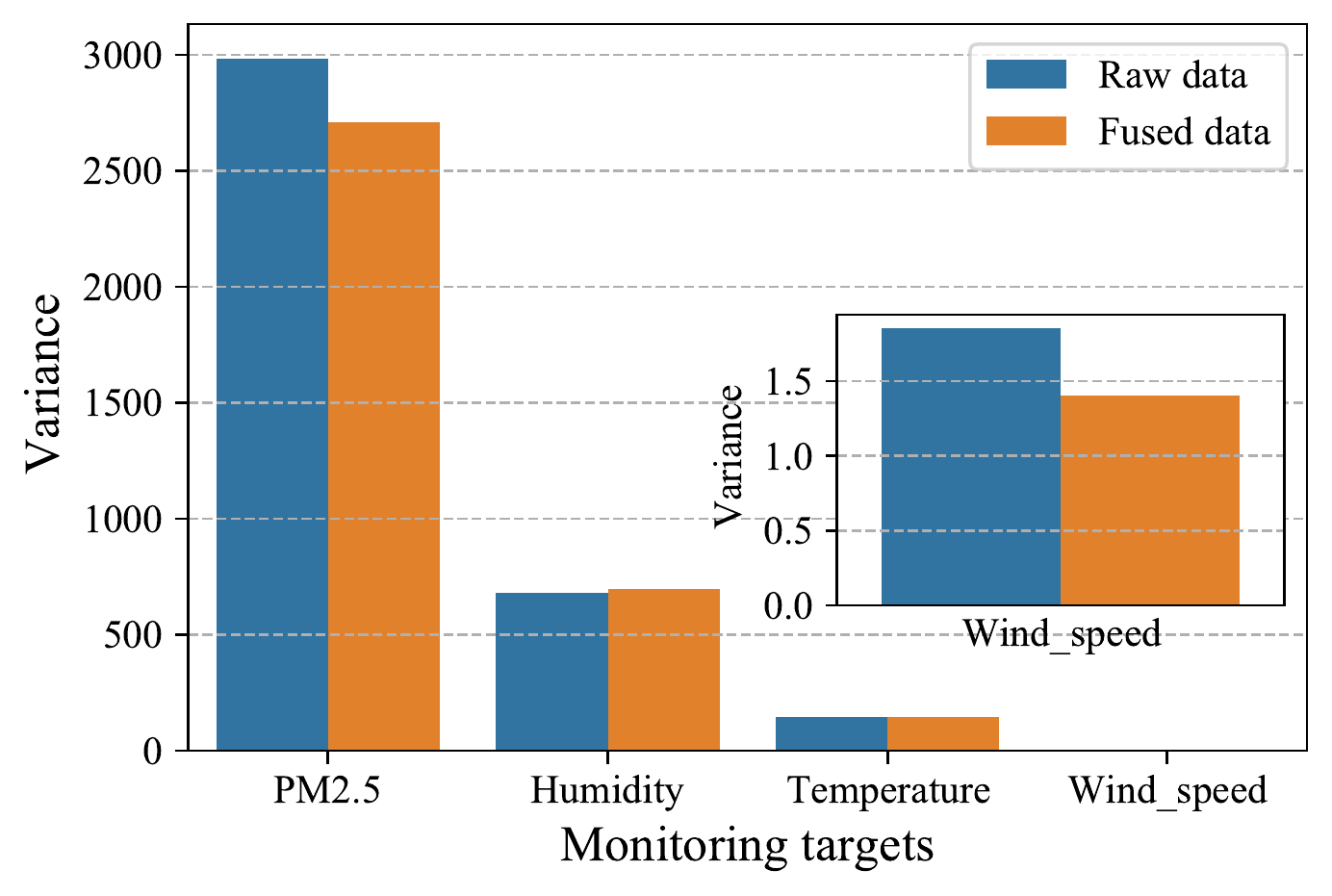}}
	\subfigure[]{
		\label{fig:subfig:9b} 
		\includegraphics[scale=0.5]{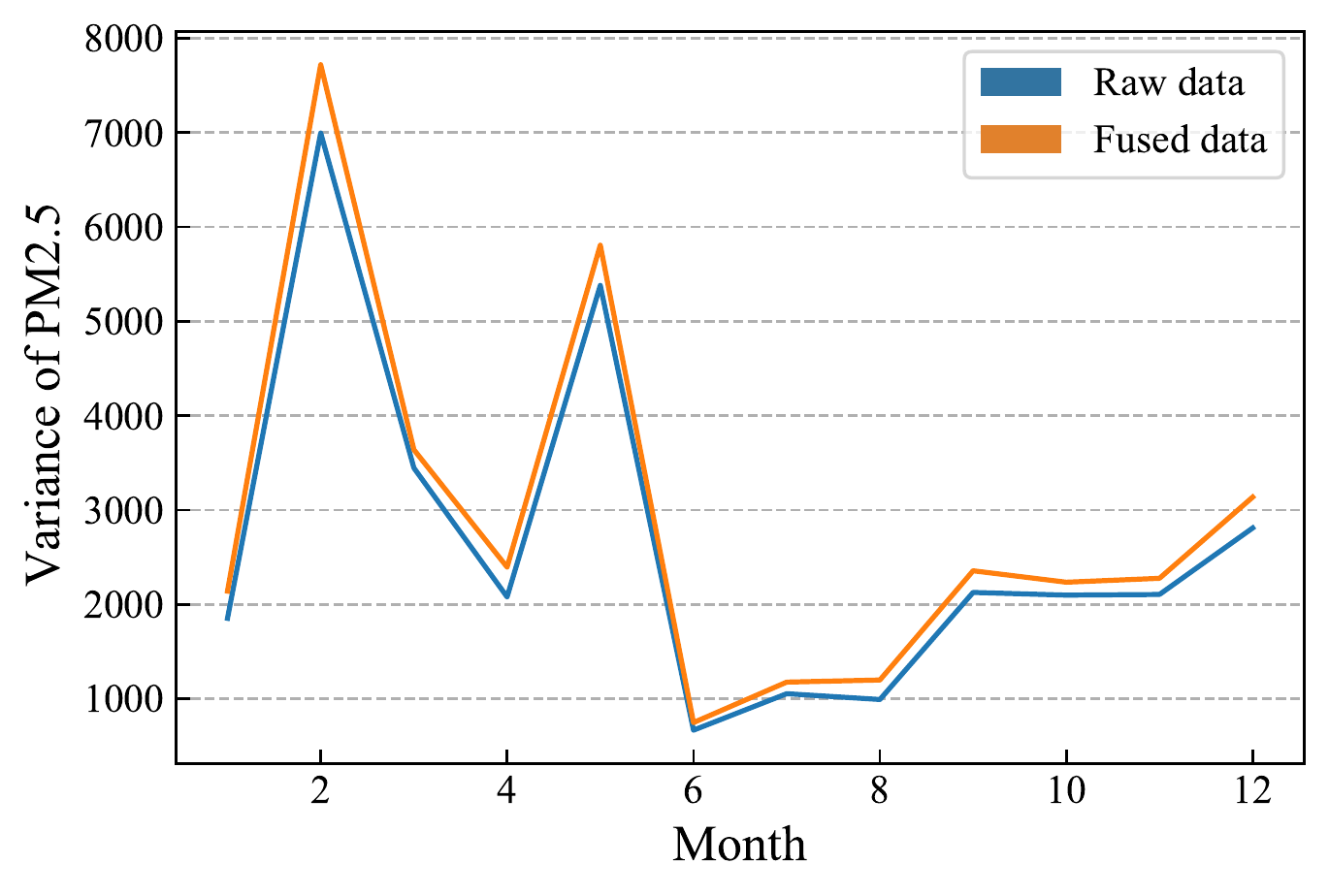}}
	\caption{The variance of the raw and fused data. (a)	Comparison on the variance of all observations. (b) Comparison on temporal distribution of the variance.} 
	\label{Figure9}
\end{figure}

First, we compared the variances of the raw and fused data, as illustrated in Figure \ref{Figure9}. The variance of the fused data achieved satisfactory consistency with the variance of the raw data. This result implies that the proposed RBF-based fusion method enables heterogeneous data to achieve a data distribution similar to that of the raw data. Furthermore, for one observation (e.g., the concentration of PM2.5), we compared the temporal distribution of the variance of the fused data with that of the raw data. The results indicate that the trend exhibited satisfactory consistency.

\begin{figure}[!ht]
	\centering
	\subfigure[]{
		\label{fig:subfig:10a} 
		\includegraphics[scale=0.5]{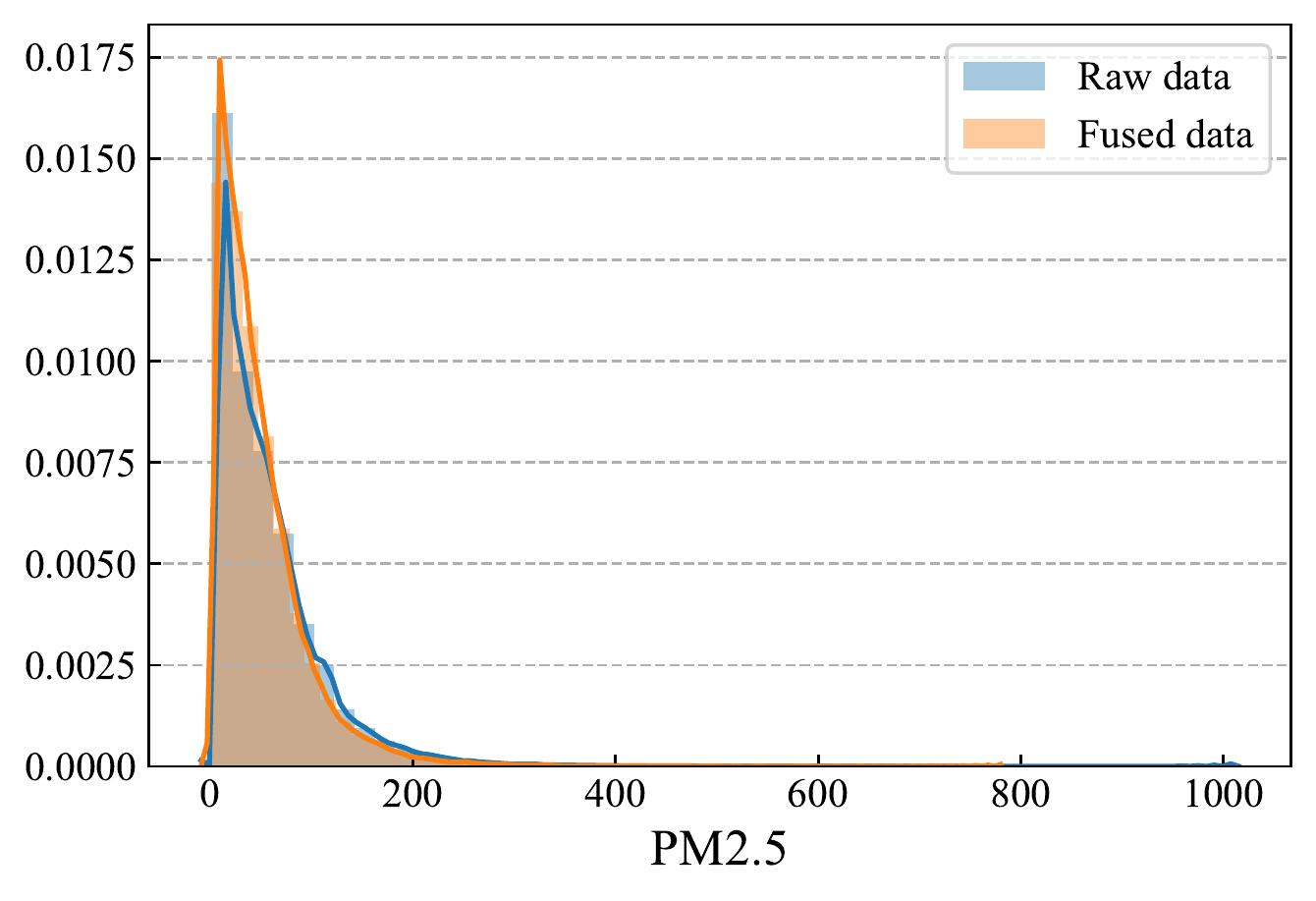}}
	\subfigure[]{
		\label{fig:subfig:10b} 
		\includegraphics[scale=0.5]{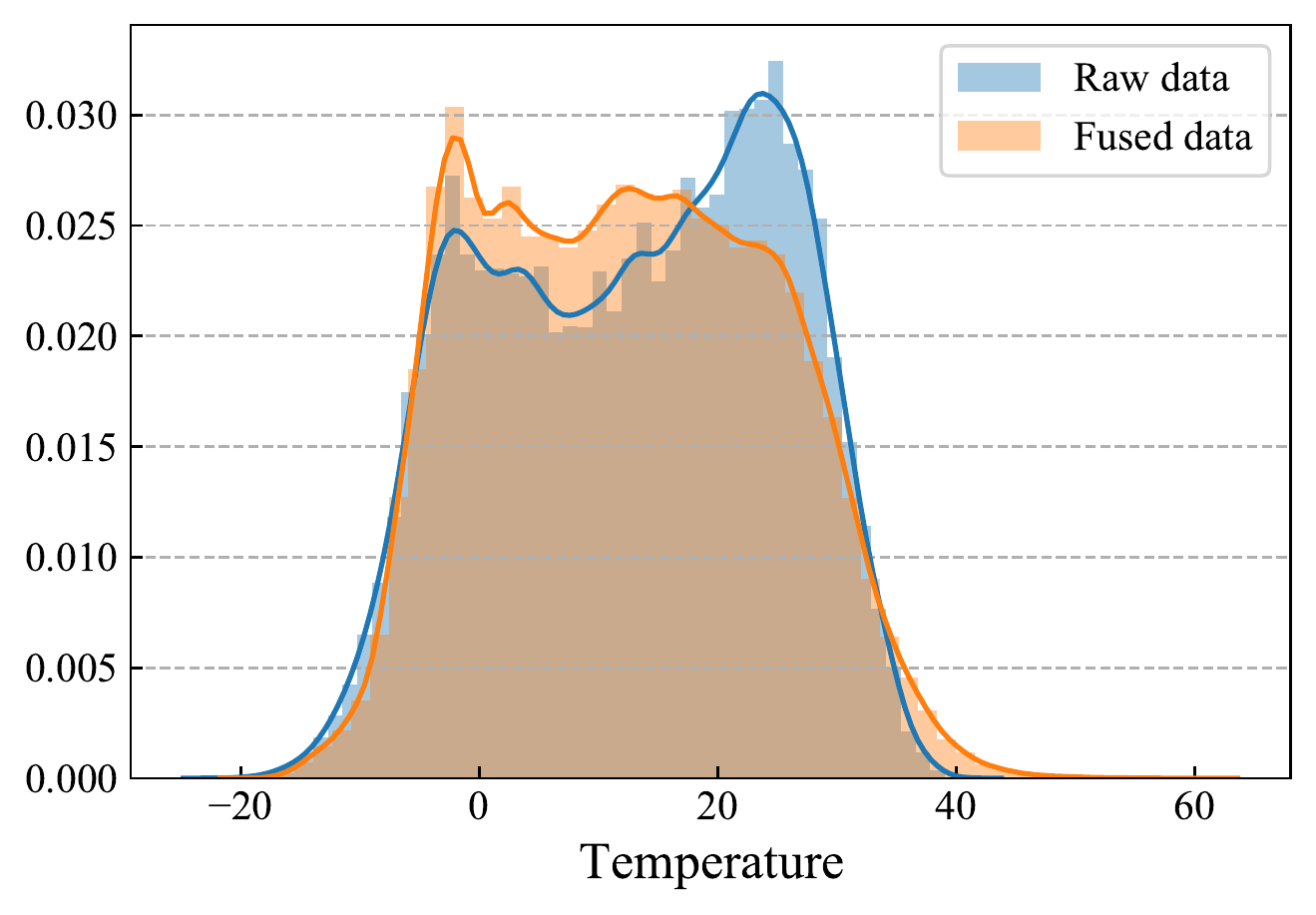}}
	\subfigure[]{
	\label{fig:subfig:10c} 
	\includegraphics[scale=0.5]{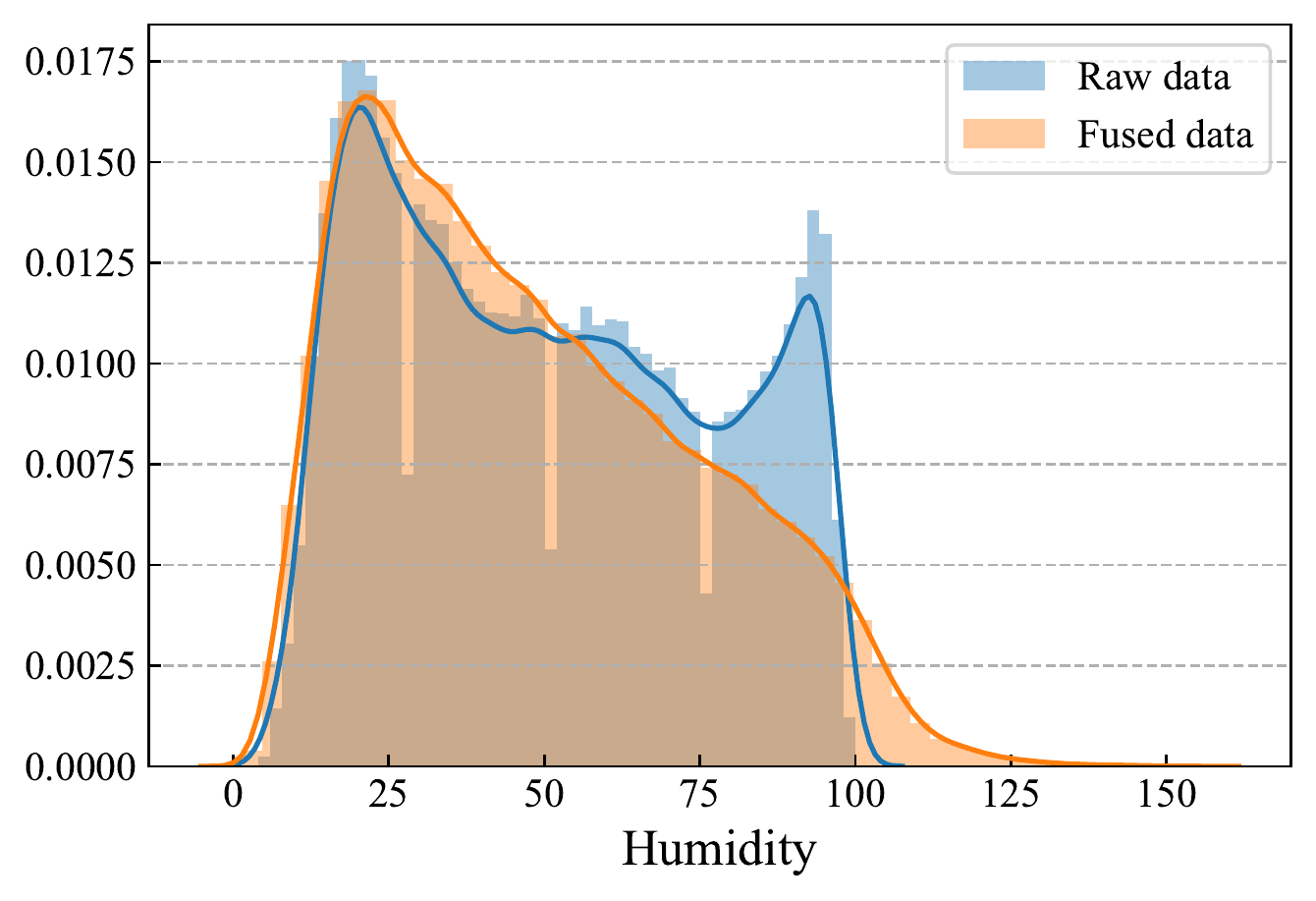}}
\subfigure[]{
	\label{fig:subfig:10d} 
	\includegraphics[scale=0.5]{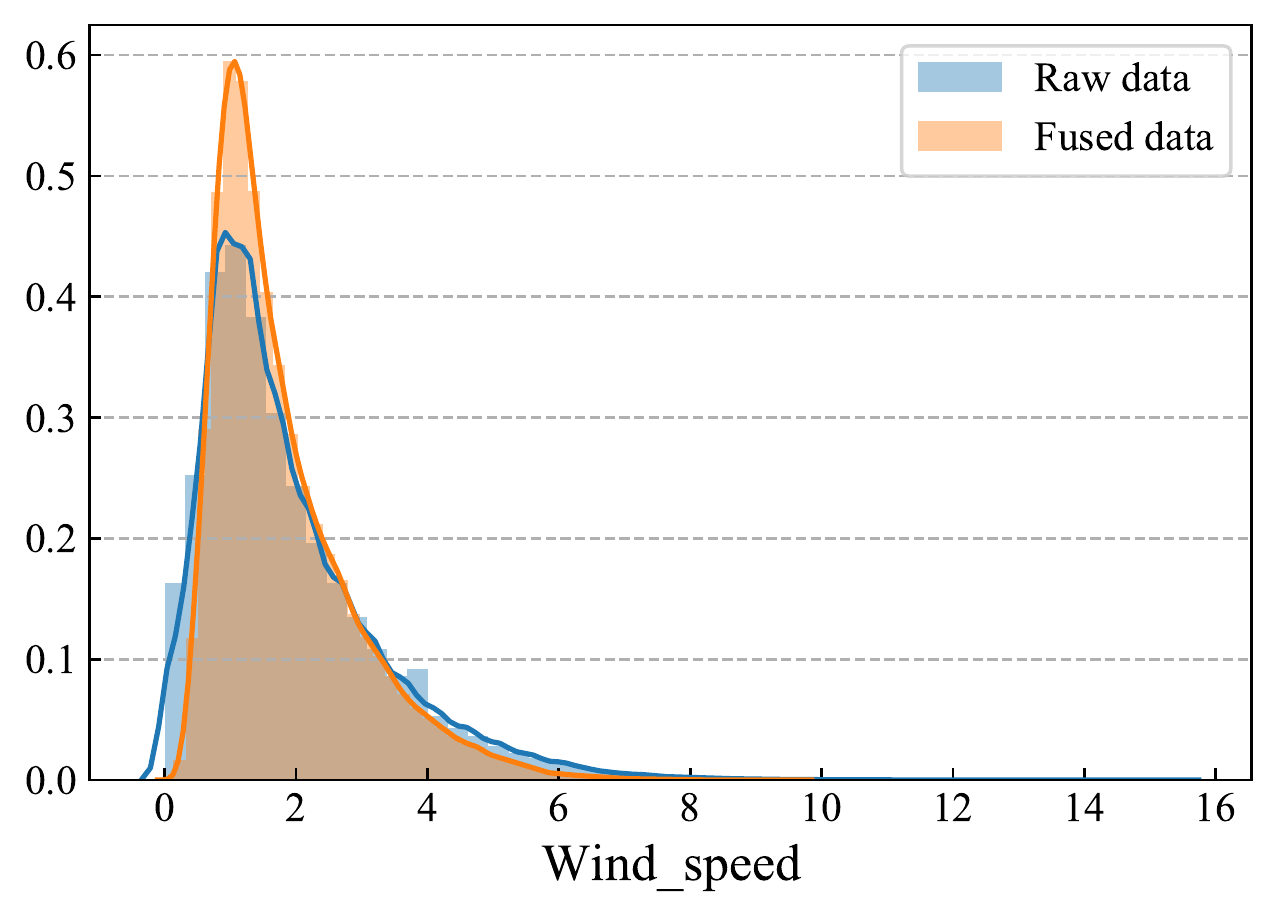}}	
	\caption{The kernel function estimates of the raw and fused data} 
	\label{Figure10}
\end{figure}

Second, we compared the kernel density estimation of the raw and fused data, as illustrated in Figure \ref{Figure10}. A kernel density estimation is an indicator for comparing the distributions of two batches of data by estimating their probability densities. The density trajectories of the raw and fused data appear to be almost superimposed, further demonstrating that the distribution of the fused data is remarkably similar to that of the raw data.

\begin{figure}[!ht]
	\centering
	\subfigure[]{
		\label{fig:subfig:11a} 
		\includegraphics[scale=0.5]{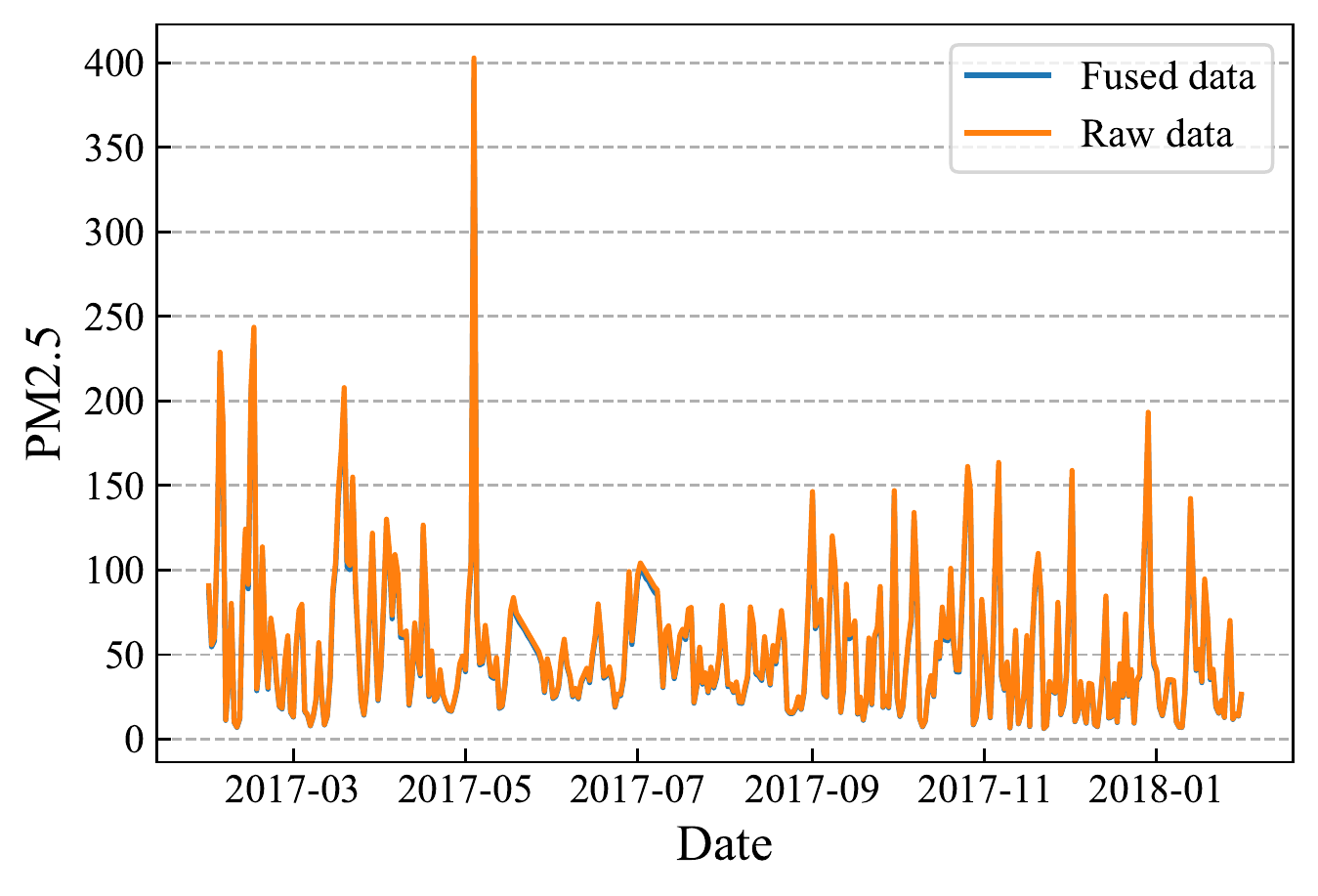}}
	\subfigure[]{
		\label{fig:subfig:11b} 
		\includegraphics[scale=0.5]{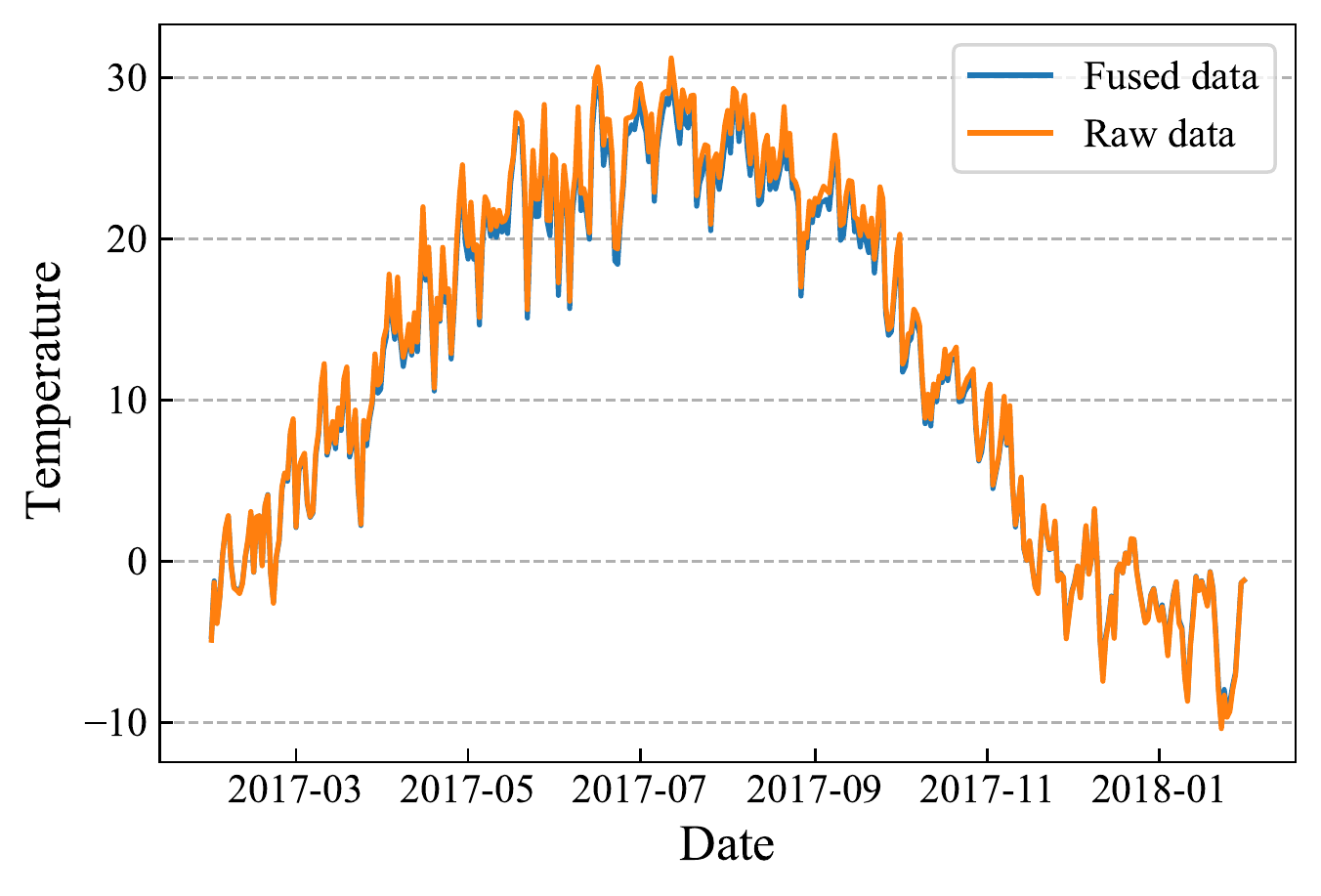}}
	\subfigure[]{
		\label{fig:subfig:11c} 
		\includegraphics[scale=0.5]{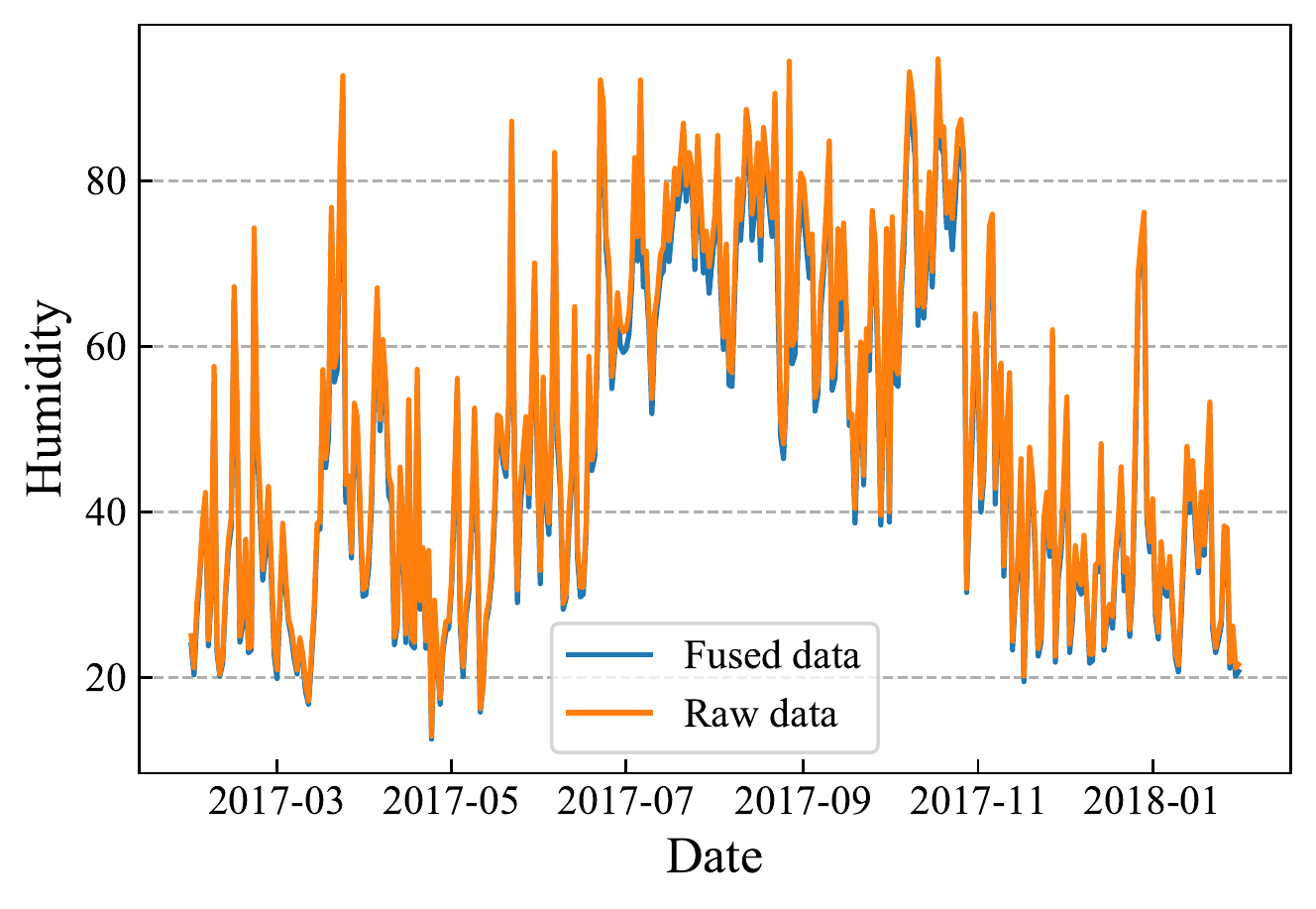}}
	\subfigure[]{
		\label{fig:subfig:11d} 
		\includegraphics[scale=0.5]{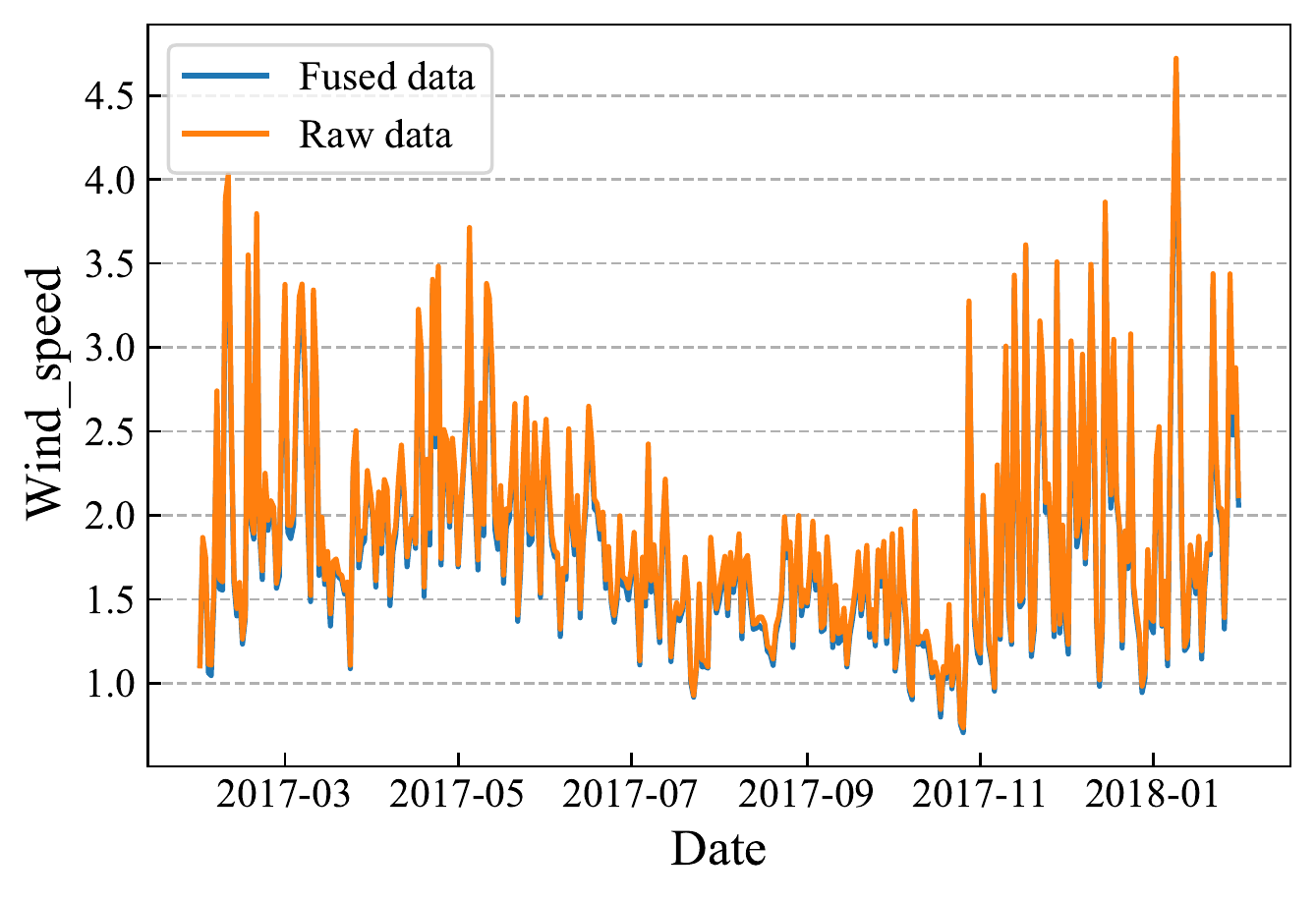}}	
	\caption{The time series distribution of the raw and fused data} 
	\label{Figure11}
\end{figure}

Third, we compared the raw and fused data in terms of their time series distributions for different monitoring targets. The time series distributions of the raw and fused data are illustrated in Figure \ref{Figure11}. Obviously, the raw and fused data for the four monitoring targets maintain temporal consistency.

\subsubsection{Effectiveness Evaluation of Data Fusion}

To further evaluate the effectiveness of the proposed RBF-based data fusion method, we randomly selected an air quality monitoring station and compared its predictions based on raw data and fused data by executing five ensemble machine learning algorithms. The machine learning algorithms used include the extra trees regressor (ETR) \cite{24}, adaptive boosted decision tree (Ada-DT) \cite{25}, random forest (RF) \cite{26}, gradient boosted decision tree (GBRT) \cite{27}, and eXtreme gradient boosting (Xgboost) \cite{28}. It should be noted that we simply aimed to perform a rough comparison between the raw and fused data, and thus the number of estimators for these ensemble machine learning algorithms were set to 200, with default values used for all remaining parameters. Here, the raw data predictions were directly obtained using data from nearby meteorological monitoring stations as features. Each data sample was divided into training and test datasets at a ratio of 8:2.

\begin{figure}[!ht]
	\centering
	\subfigure[]{
		\label{fig:subfig:12a} 
		\includegraphics[scale=0.5]{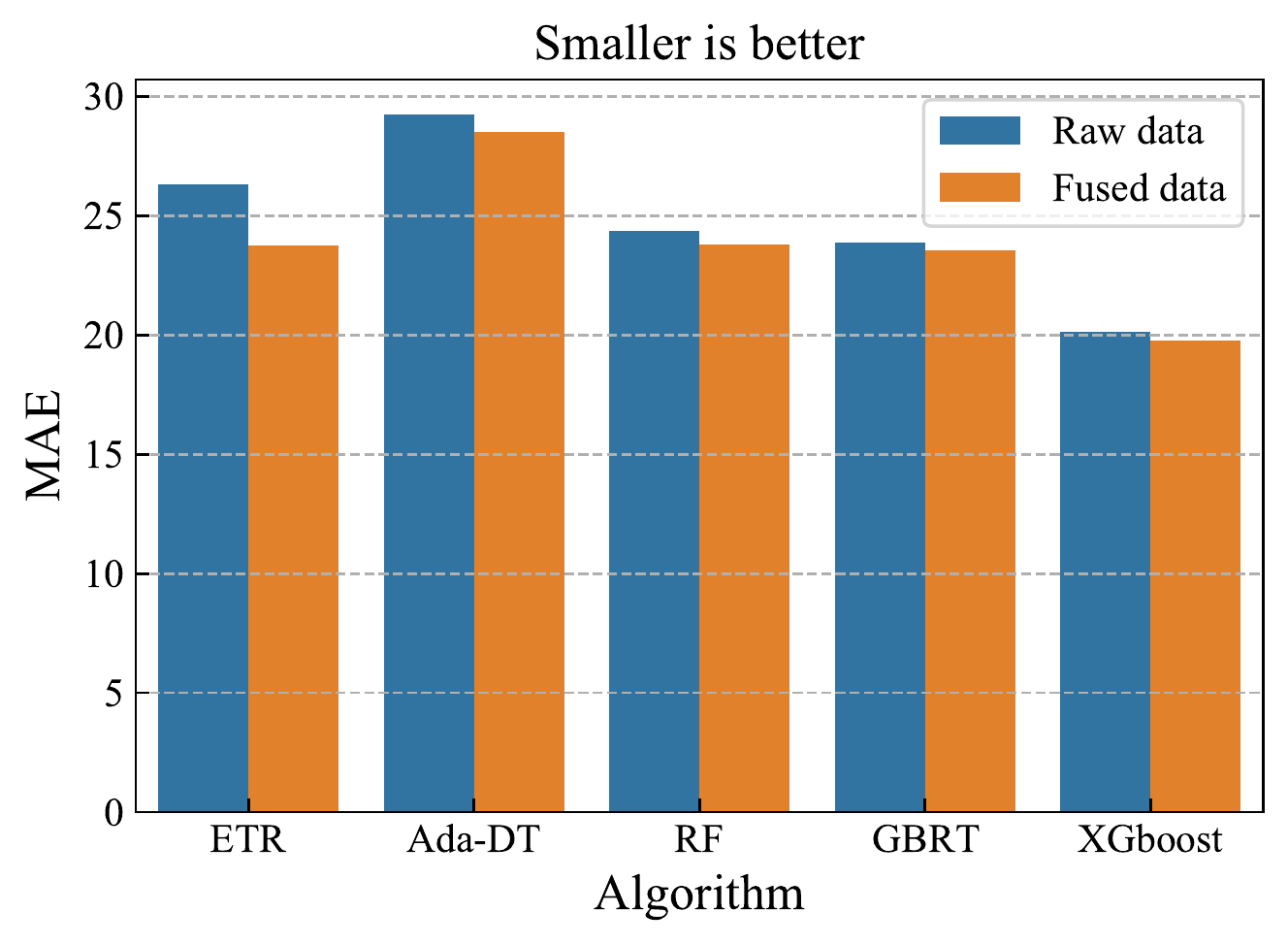}}
	\subfigure[]{
		\label{fig:subfig:12b} 
		\includegraphics[scale=0.5]{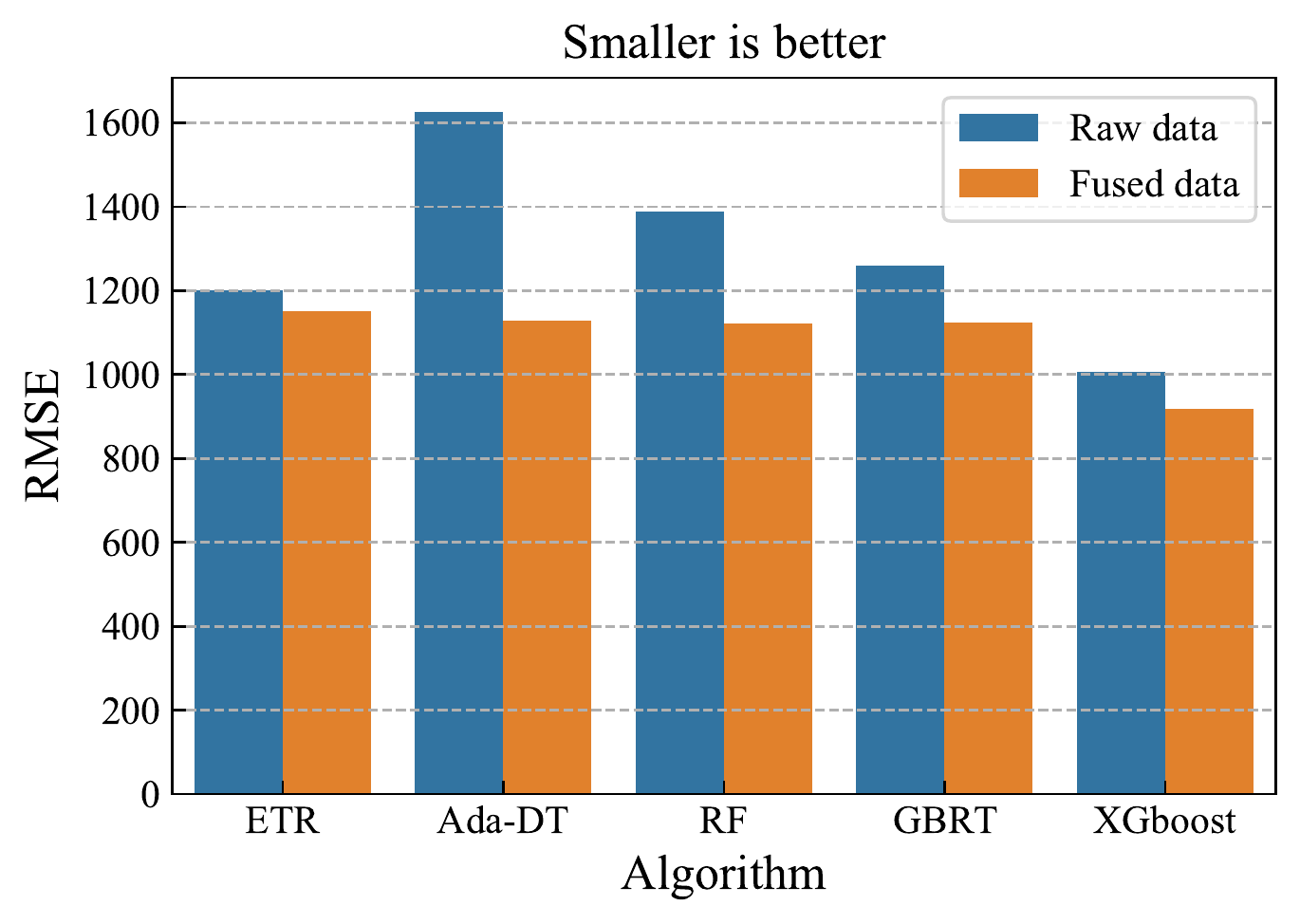}}
	\subfigure[]{
		\label{fig:subfig:12c} 
		\includegraphics[scale=0.5]{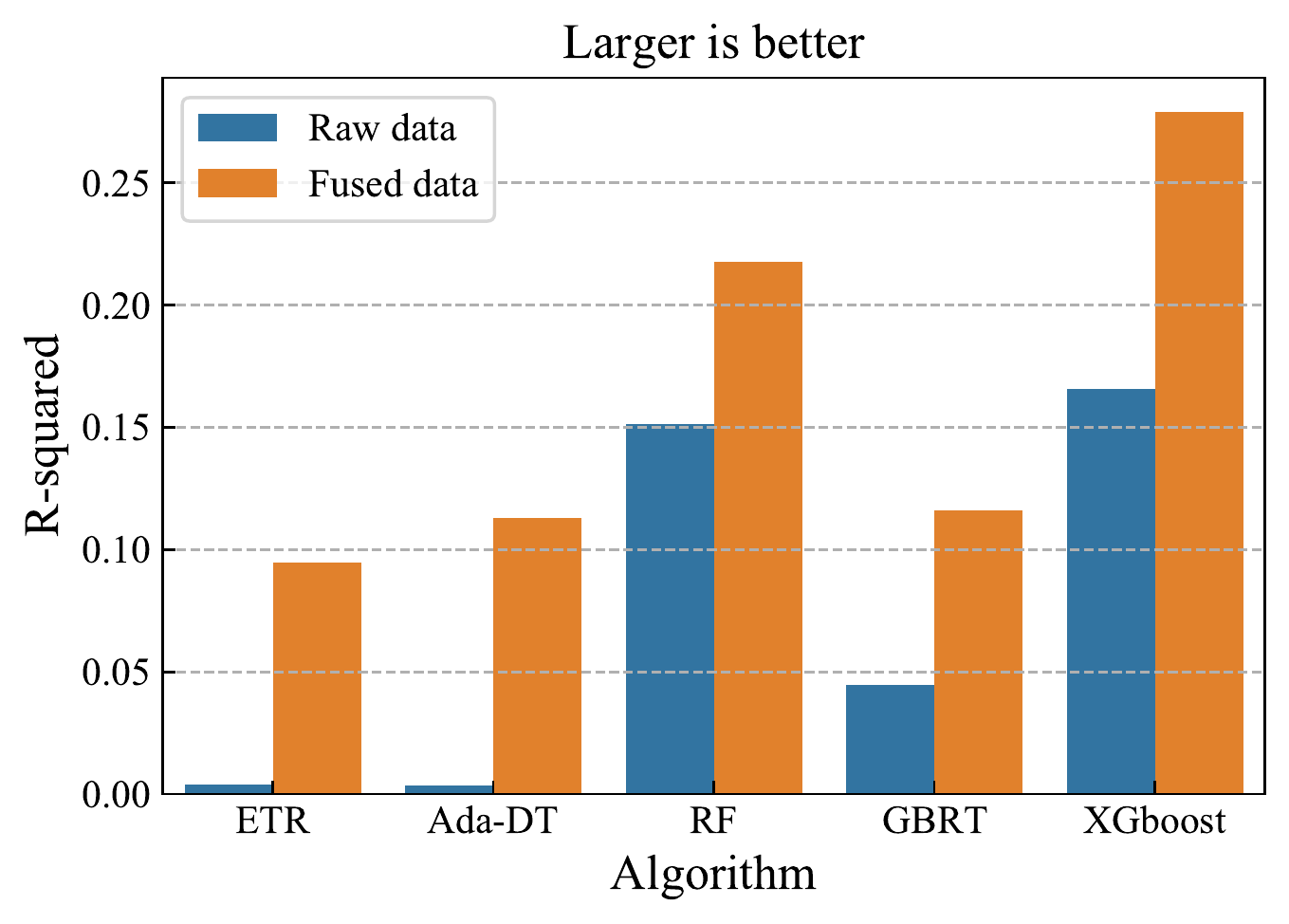}}
	\subfigure[]{
		\label{fig:subfig:12d} 
		\includegraphics[scale=0.5]{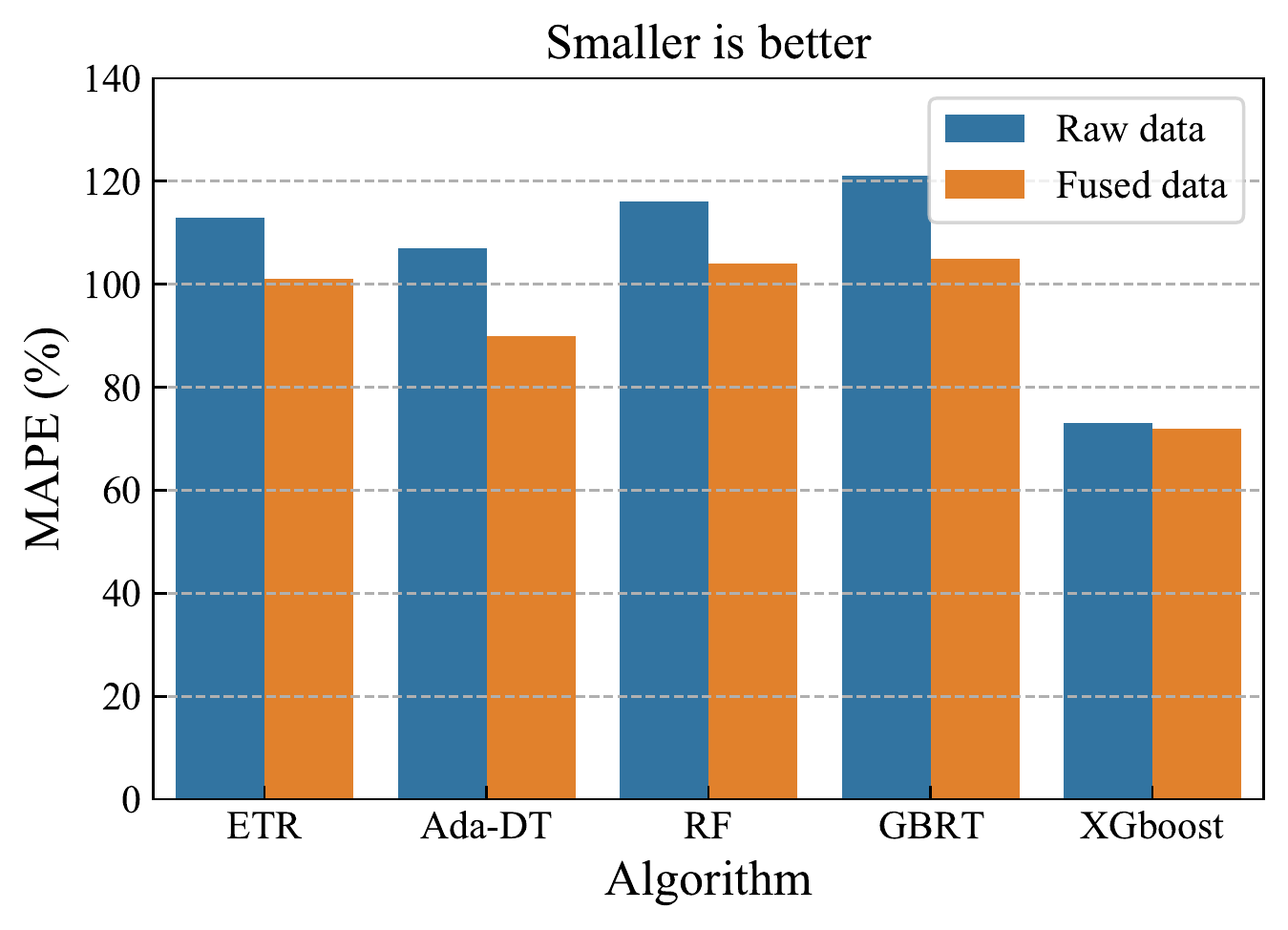}}	
	\caption{Comparison of the performance between fused data and raw data in the traditional machine learning model. (a) MAE. (b) RMSE. (c) $\mathrm{R}^{2}$. (d) MAPE} 
	\label{Figure12}
\end{figure}

We adopted four common metrics to evaluate these five models’ performances: (1) mean absolute error (MAE), (2) mean absolute percentage error (MAPE), (3) root mean square error (RMSE), and (4) R-squared ($\mathrm{R}^{2}$); see Eqs. (\ref{equ:eq18}), (\ref{equ:eq19}), (\ref{equ:eq20}), and (\ref{equ:eq21}). For the MAE, MAPE, and RMSE, a smaller metric value indicates a better performance by the prediction model. For $\mathrm{R}^{2}$, in the range of 0 to 1, a larger metric indicates a better performance by the prediction model. All metrics results of the above five ensemble machine learning algorithms are presented in Table \ref{table:tab1} and Figure \ref{Figure12}.

\begin{equation}
\label{equ:eq18}
\mathrm{MAE}=\frac{1}{\eta} \sum_{t=1}^{\eta}\left|\chi_{t}-\hat{\chi}_{t}\right|
\end{equation}

\begin{equation}
\label{equ:eq19}
R M S E=\sqrt{\frac{1}{\eta} \sum_{t}\left(\chi_{t}-\hat{\chi}_{t}\right)^{2}}
\end{equation}

\begin{equation}
\label{equ:eq20}
M A P E=\frac{100 \%}{\eta} \sum_{t=1}^{\eta}\left|\frac{\hat{\chi}_{t}-\chi_{t}}{\chi_{t}}\right|
\end{equation}

\begin{equation}
\label{equ:eq21}
R^{2}=1-\frac{\sum\left(\chi_{t}-\hat{\chi}_{t}\right)^{2}}{\sum\left(\chi_{t}-\hat{\chi}_{t}\right)^{2}}
\end{equation}
where $\left\{\hat{\chi}_{t}\right\}$ and $\left\{\chi_{t}\right\}$ denote the predicted results and ground truth, respectively. $\eta$ denotes the total number of all predicted values.

\begin{table}[]
   \renewcommand{\arraystretch}{1.3}
	\caption{Comparison of the performance when using the fused data and raw data in the traditional machine learning model}
	\label{table:tab1}
	\centering
	\begin{tabular}{cccccc}
		\hline
		Algorithm                     & Data                & MAE            & RMSE             & $\mathrm{R}^{2}$              & MAPE           \\ \hline
		\multirow{2}{*}{ETR} & Raw data            & 26.33          & 1201.10          & 0.0037          & 113\%          \\
		& \textbf{Fused data} & \textbf{23.74} & \textbf{1151.63} & \textbf{0.0947} & \textbf{101\%} \\
		\multirow{2}{*}{Ada-DT}       & Raw data            & 29.25          & 1626.05          & 0.0034          & 107\%          \\
		& \textbf{Fused data} & \textbf{28.53} & \textbf{1128.20} & \textbf{0.1131} & \textbf{90\%}  \\
		\multirow{2}{*}{RF}           & Raw data            & 24.37          & 1388.08          & 0.1513          & 116\%          \\
		& \textbf{Fused data} & \textbf{23.78} & \textbf{1122.56} & \textbf{0.2175} & \textbf{104\%} \\
		\multirow{2}{*}{GBRT}         & Raw data            & 23.87          & 1259.66          & 0.0448          & 121\%          \\
		& \textbf{Fused data} & \textbf{23.55} & \textbf{1124.43} & \textbf{0.1161} & \textbf{105\%} \\
		\multirow{2}{*}{Xgboost}      & Raw data            & 20.15          & 1005.76          & 0.1657          & 73\%           \\
		& \textbf{Fused data} & \textbf{19.77} & \textbf{917.21}  & \textbf{0.2790} & \textbf{72\%}  \\ \hline
	\end{tabular}
\end{table}

As presented in Table \ref{table:tab1} and Figure \ref{Figure12}, for all five machine learning algorithms, all the fused data have smaller MAE, RMSE, and MAPE values and larger $\mathrm{R}^{2}$ values than those of the raw data. The above results indicate that the performances of the prediction models based on fused data are much better than those based on raw data. The effectiveness of the data derived from the proposed RBF-based data fusion method for developing predictive models is thus demonstrated.

\subsection{Spatiotemporal Predication Using Graph Neural Networks}
\subsubsection{Construction of Spatially and Temporally Correlated Data}

\begin{figure}[!ht]
	\centering
	\subfigure[]{
		\label{fig:subfig:13a} 
		\includegraphics[scale=0.4]{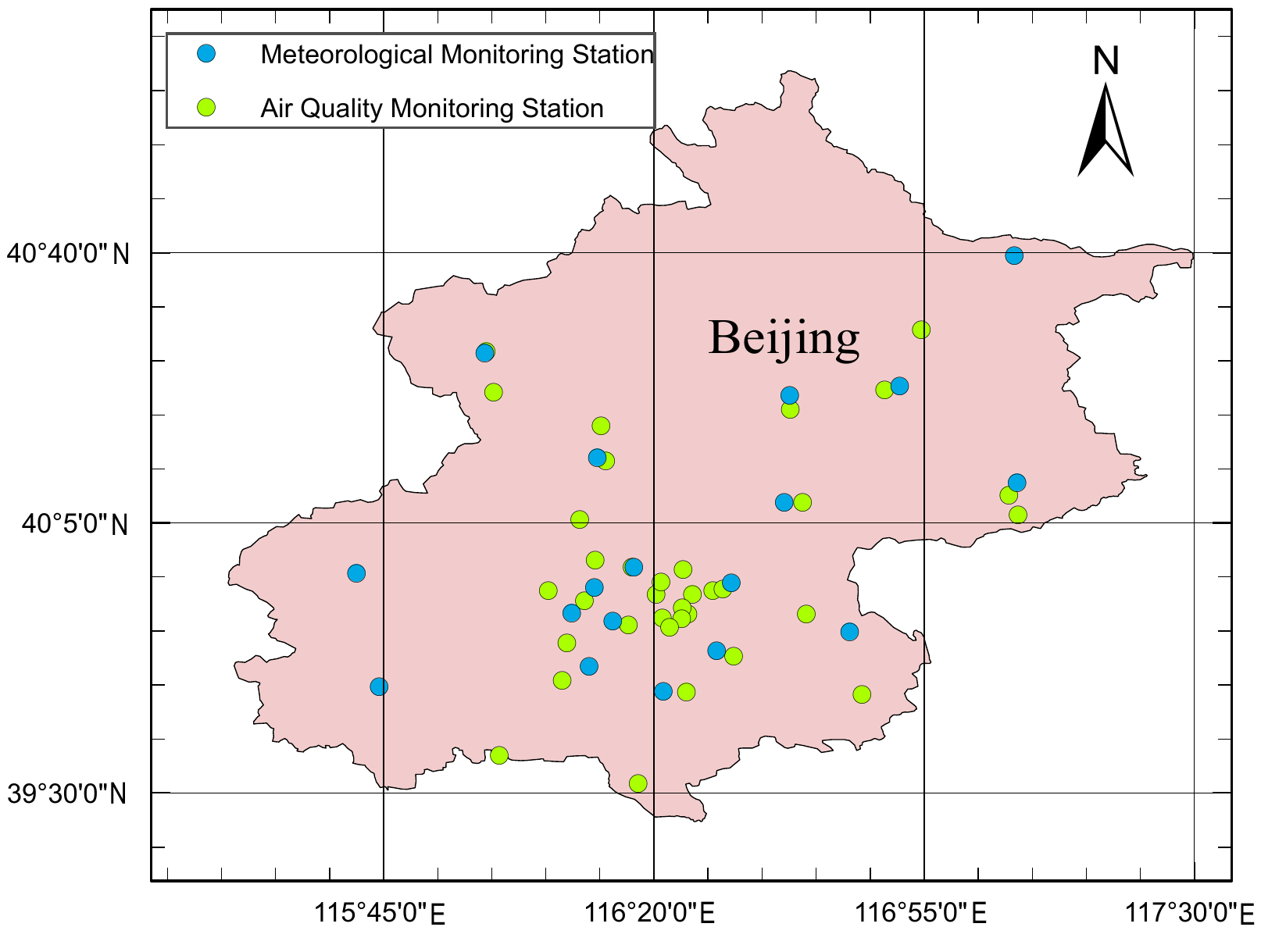}}
	\subfigure[]{
		\label{fig:subfig:13b} 
		\includegraphics[scale=0.153]{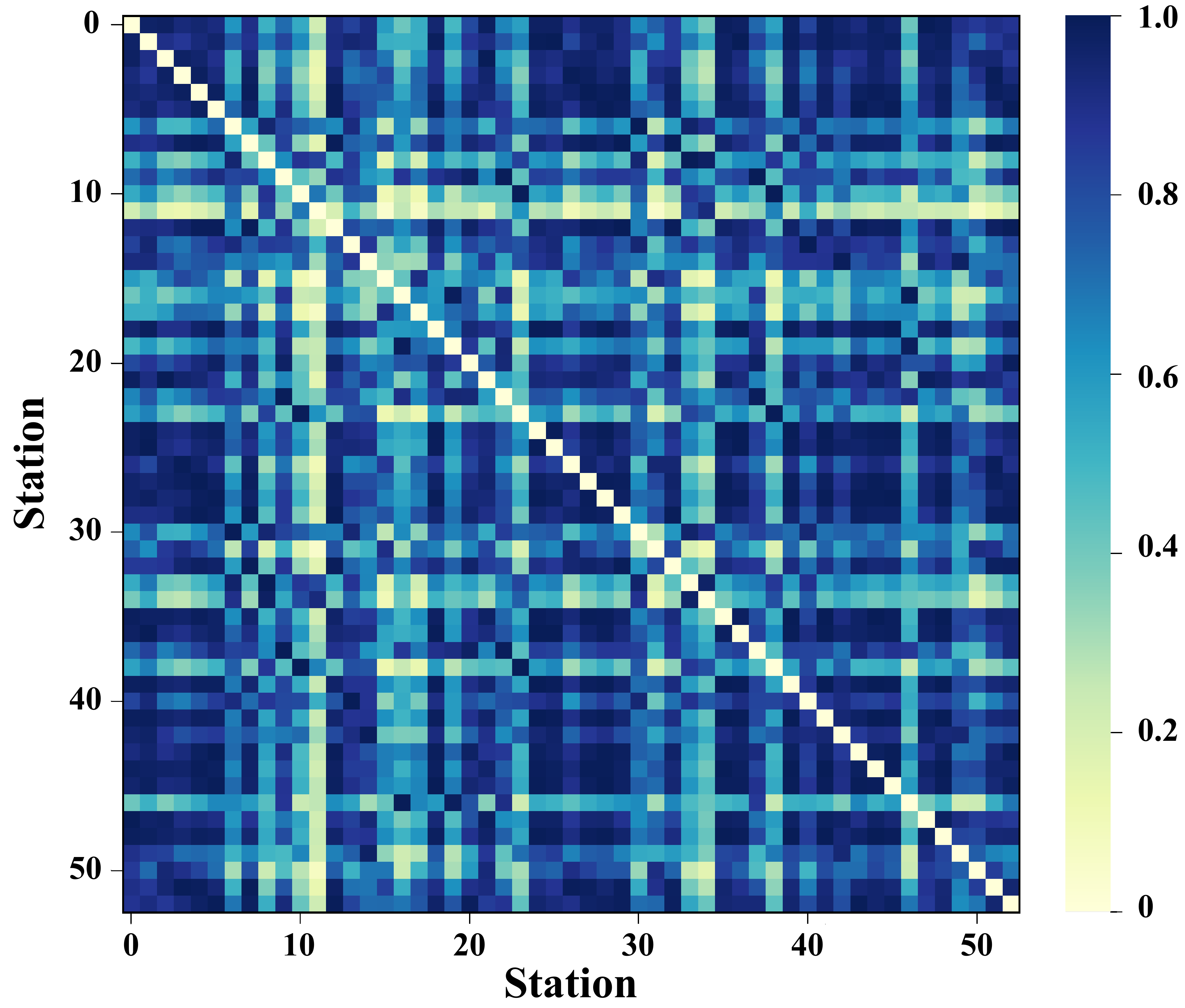}}
	\caption{(a) The layout of all monitoring stations. (b) Heat map of the weighted adjacency matrix $A_{w}$} 
	\label{Figure13}
\end{figure}

Data construction consists of (1) the construction of spatially correlated data and (2) the construction of temporally correlated data. For the spatially correlated data, to construct a weighted fully connected graph, we connected the 53 monitoring stations to each other to assemble a weighted matrix $A_{w}$ with a size of $53 \times 53$ that represents the spatial relationships among multiple monitoring stations with a graphical structure. The values in the matrix denote the similarity between the monitoring stations in the study area. The visualization of $A_{w}$ is illustrated in Figure \ref{fig:subfig:13b}, where the darker the color is, the greater the correlation. For the temporally correlated data, we defined the data interval as once every hour. Consequently, the size of the constructed sequence data is $8784 \times 53 \times 4$, where 8784 is the total length of the time series with hourly intervals, 53 is the total number of monitoring stations scattered across the entire study area, and 4 is the sum of the numbers of monitoring targets for all monitoring sources. Each value of the constructed sequence data represents the change in each data point over time. Furthermore, we exploited the min-max normalization method to scale the values in the range of $[0, 1]$. Finally, 60\% of the data were utilized for training, 20\% of the data were utilized for validation, and the remaining 20\% were utilized for testing.

\subsubsection{Experimental Settings}

The experimental settings included (1) the implementation and testing environment, (2) the determination of model parameters, and (3) the baseline methods adopted for performance evaluation purposes.

We implemented the proposed deep learning model based on the PyTorch framework \cite{29}. The training process of the model was performed on a workstation computer equipped with a Quadro P6000 GPU with 24 GB of GPU memory, an Intel Xeon Gold 5118 CPU, and 128 GB of RAM.

The detailed hyperparameter settings for the model were as follows. The first step is to set the model parameters. Both the size of the graph convolution kernel in the GCN and the size of the time convolution kernel in the GCNN were 3. In addition, we added a dropout layer with a rate of 0.3. The next step is to set the training parameters of the model. We set the learning rate to 0.001, the batch size to 32, and the training period to 350, and we employed the Adam optimizer to minimize the loss function during the training process \cite{30}. We utilized 12 hours as the historical time step $P$ for predicting the concentration of PM2.5 in the next step $Q = 3$ (i.e., 3 hours). Furthermore, we adopted two common metrics, i.e., the MAE and RMSE, to evaluate the performance of the model.

To comparatively evaluate the performance of the proposed method, we employed three other baseline deep learning methods: (1) LSTM, a variant of the recurrent neural network (RNN) that is able to analyze time-series data \cite{31}; (2) GRU, a variant of the RNN that is able to analyze time-series data \cite{32}; and (3) T-GCN, a model which fuses the spatial and temporal correlations of a single observation to perform prediction \cite{33}. Furthermore, we compared the performance of the STGCN model based on different numbers of monitored targets, expressed as STCGN-K1, STGCN-K2, STGCN-K3, and STCGN-K4. K1 refers to utilizing the predictive targets (i.e., concentrations of PM2.5) as inputs. K2 refers to utilizing the predictive targets and one sampling observation as inputs. K3 refers to utilizing the predictive targets and two sampling observations as inputs. K4 refers to utilizing all observations in the dataset as inputs.

\subsubsection{Predicted Results}

\begin{figure}[!ht]
	\centering
	\subfigure[MAE]{
		\label{fig:subfig:14a} 
		\includegraphics[scale=0.5]{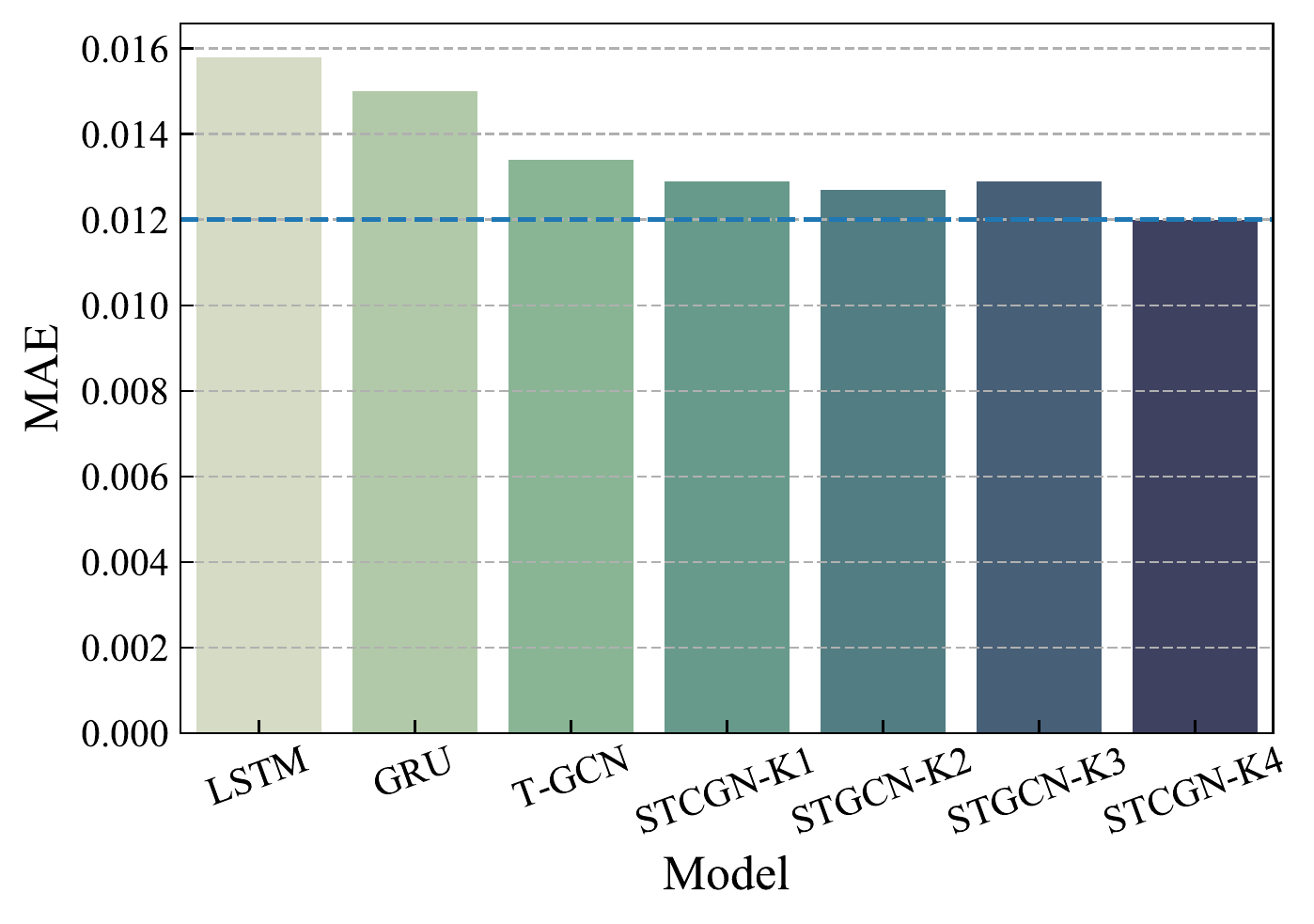}}
	\subfigure[RMSE]{
		\label{fig:subfig:14b} 
		\includegraphics[scale=0.5]{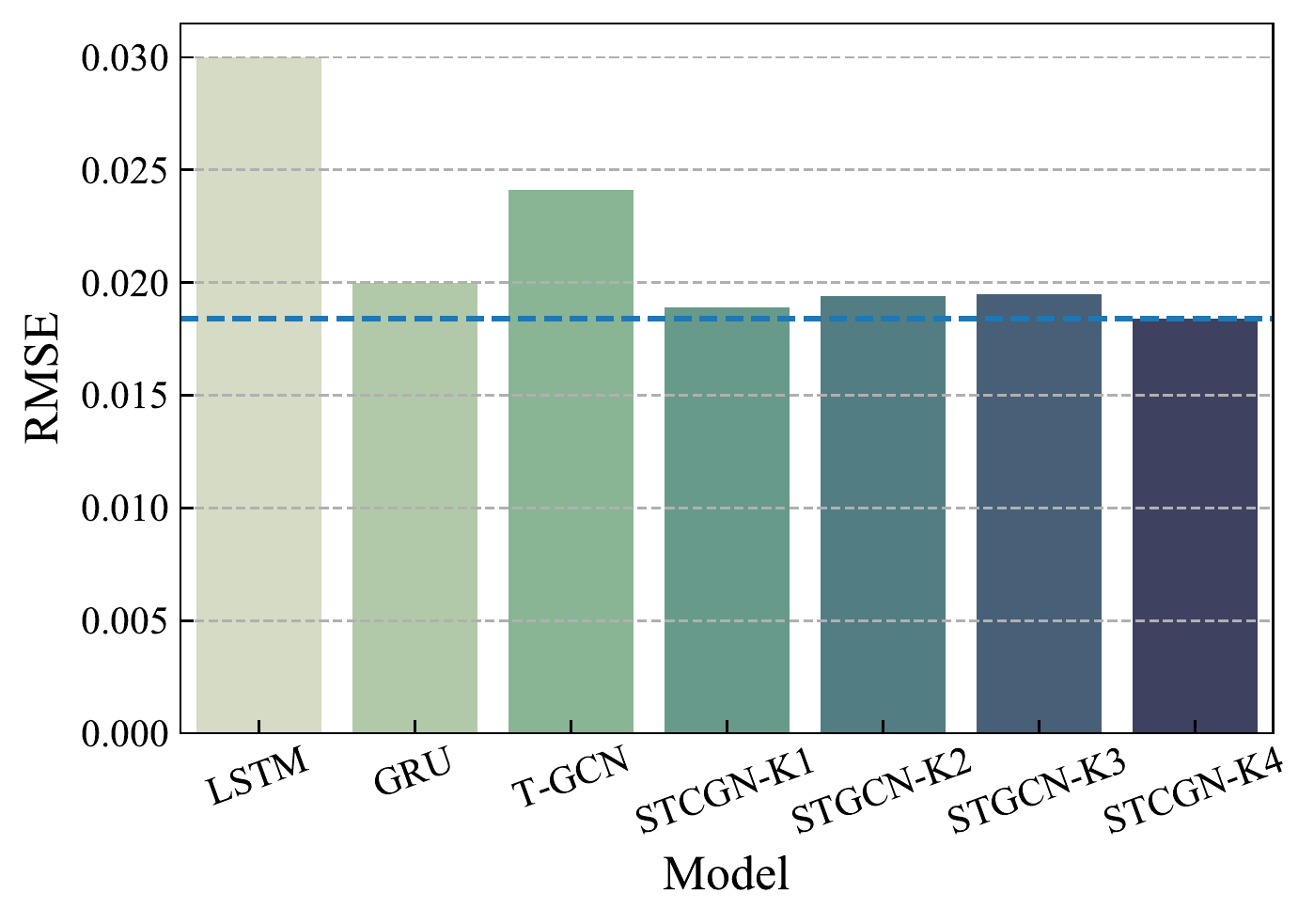}}
	\caption{Performance comparison of the three baseline models and the employed STGCN model} 
	\label{Figure14}
\end{figure}

We compared the performances of the aforementioned models on the constructed data; for details on the data construction process, see Subsection \ref{sec:2.5:Dataconstruction}. The results are presented in Table \ref{table:tab2} and Figure \ref{Figure14}.

\begin{table}[]
    \renewcommand{\arraystretch}{1.3}
	\caption{Comparison of the performance of the predictions}
	\label{table:tab2}
	\centering
	\begin{tabular}{ccc}
		\hline
		Model             & MAE             & RMSE            \\ \hline
		LSTM              & 0.0158          & 0.0300          \\
		GRU               & 0.0150          & 0.0200          \\
		T-GCN             & 0.0134          & 0.0241          \\
		STCGN-K1          & 0.0129          & 0.0189          \\
		STGCN-K2          & 0.0127          & 0.0194          \\
		STGCN-K3          & 0.0129          & 0.0195          \\
		\textbf{STCGN-K4} & \textbf{0.0120} & \textbf{0.0184} \\ \hline
	\end{tabular}
\end{table}

As illustrated in Table \ref{table:tab2} and Figure \ref{Figure14}, the STGCN model achieves the best performance when compared to those of all the baseline models. Among these models, LSTM and GRU perform predictions utilizing time series data for a single observation (i.e., concentration of PM2.5) of a single monitoring station as input. Both the T-GCN and ST-GCN employ graph convolution to predict time series for the constructed graph-structured data, i.e., they both take spatiotemporal correlations into account. Their MAE values are greater than those of the traditional LSTM and GRU. The STGCN is capable of performing predictions utilizing multiple observations (i.e., external factors that influence air quality) as inputs, while the T-GCN performs predictions based on the spatiotemporal correlation of a single piece of information (i.e., air quality itself). The STGCN obtained the best performance. Furthermore, among the four STGCN models, the best performance was obtained by STGCN-K4 because its input contains all the external information about the fusion matrix, i.e., weather, temperature, and wind speed; the more information that is fused, the better the model performs. However, STGCN-K3 did not perform as well as STGCN-K2, indicating that the prediction results are presumably related to combinations of the different observations.

\begin{figure}[!ht]
	\centering
	\subfigure[]{
		\label{fig:subfig:15a} 
		\includegraphics[scale=0.5]{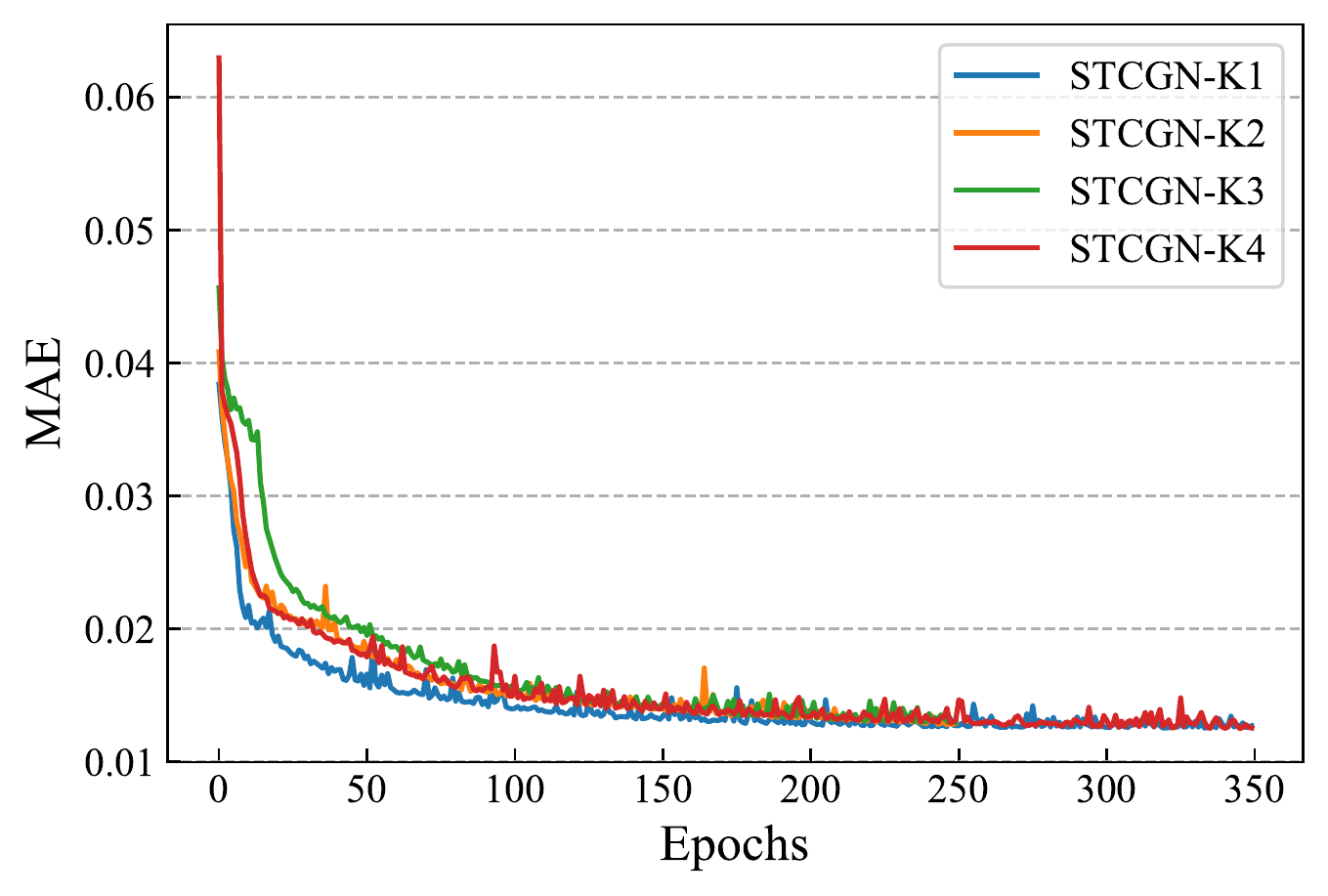}}
	\subfigure[]{
		\label{fig:subfig:15b} 
		\includegraphics[scale=0.5]{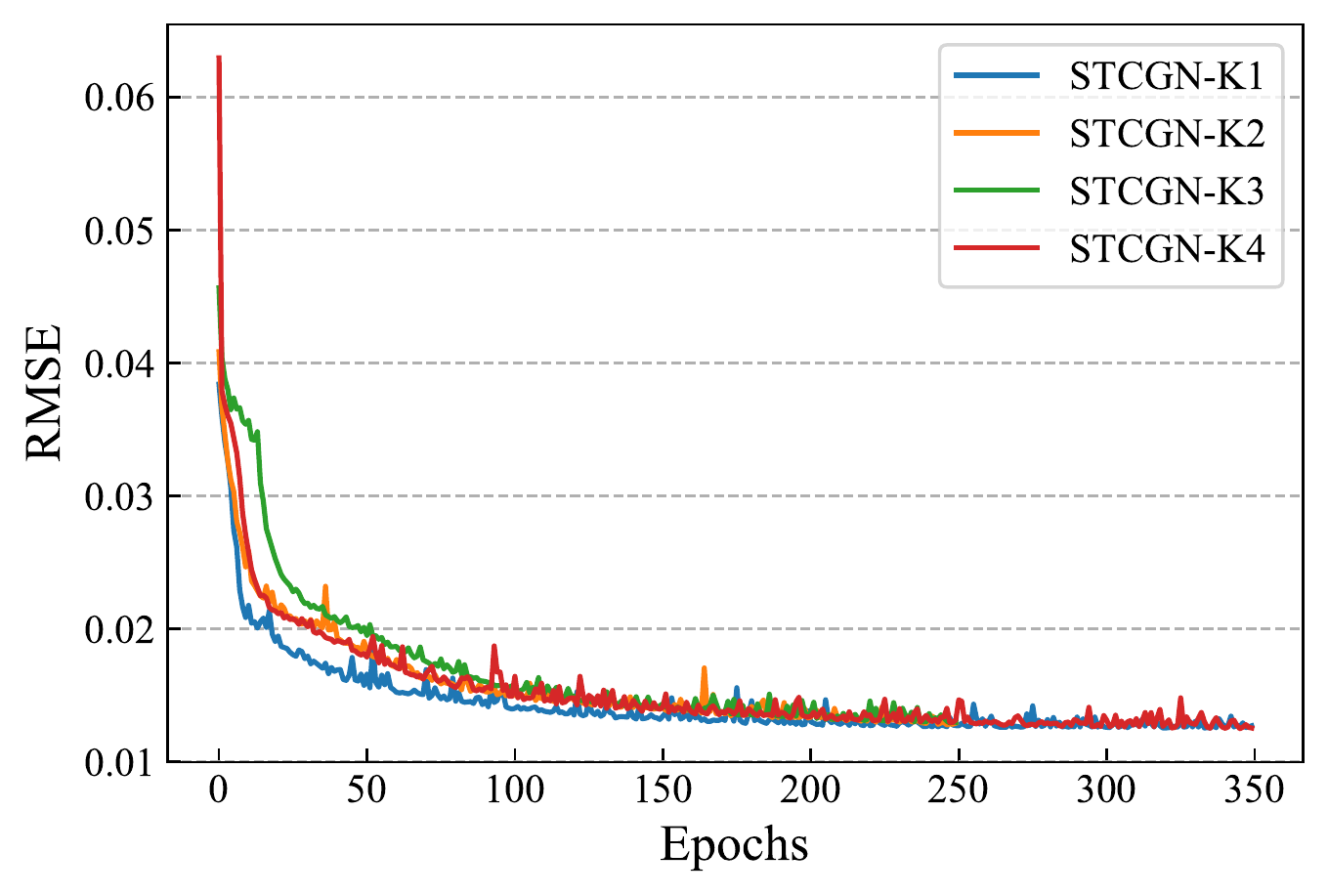}}
	\caption{(a) RMSE versus the epochs. (b) MAE versus the epochs} 
	\label{Figure15}
\end{figure}

To further investigate the effect of different amounts of observations on the model prediction results, we plotted the RMSE and MAE curves of the four STGCN models. As illustrated in Figure \ref{Figure15}, the performances of these STGCN models are similar. 

We further investigated the influences that different combinations of observations have on the prediction results. For the four monitoring targets of the utilized dataset, PM denotes the concentration of PM2.5, HU denotes the humidity, TE denotes the temperature, and WS denotes the wind speed. Different combinations of observations were fed into the STGCN model, and the prediction results are listed in Table \ref{table:tab3}.

\begin{table}[]
    \renewcommand{\arraystretch}{1.3}
	\caption{Performance comparison of different combinations of observations}
	\label{table:tab3}
	\centering
	\begin{tabular}{ccc}
		\hline
		Model              & MAE             & RMSE            \\ \hline
		STGCN-PM-HU        & 0.0127          & 0.0194          \\
		STGCN-PM-TE        & 0.0123          & 0.0189          \\
		STGCN-PM-WS        & 0.0124          & 0.0191          \\
		STGCN-PM-HU-TE     & 0.0129          & 0.0195          \\
		STGCN-PM-HU-WS     & 0.0126          & 0.0190          \\
		STGCN-PM-TE-WS     & 0.0131          & 0.0193          \\
		\textbf{STGCN-ALL} & \textbf{0.0120} & \textbf{0.0184} \\ \hline
	\end{tabular}
\end{table}

\begin{figure}[!ht]
	\centering
	\subfigure[MAE]{
		\label{fig:subfig:16a} 
		\includegraphics[scale=0.5]{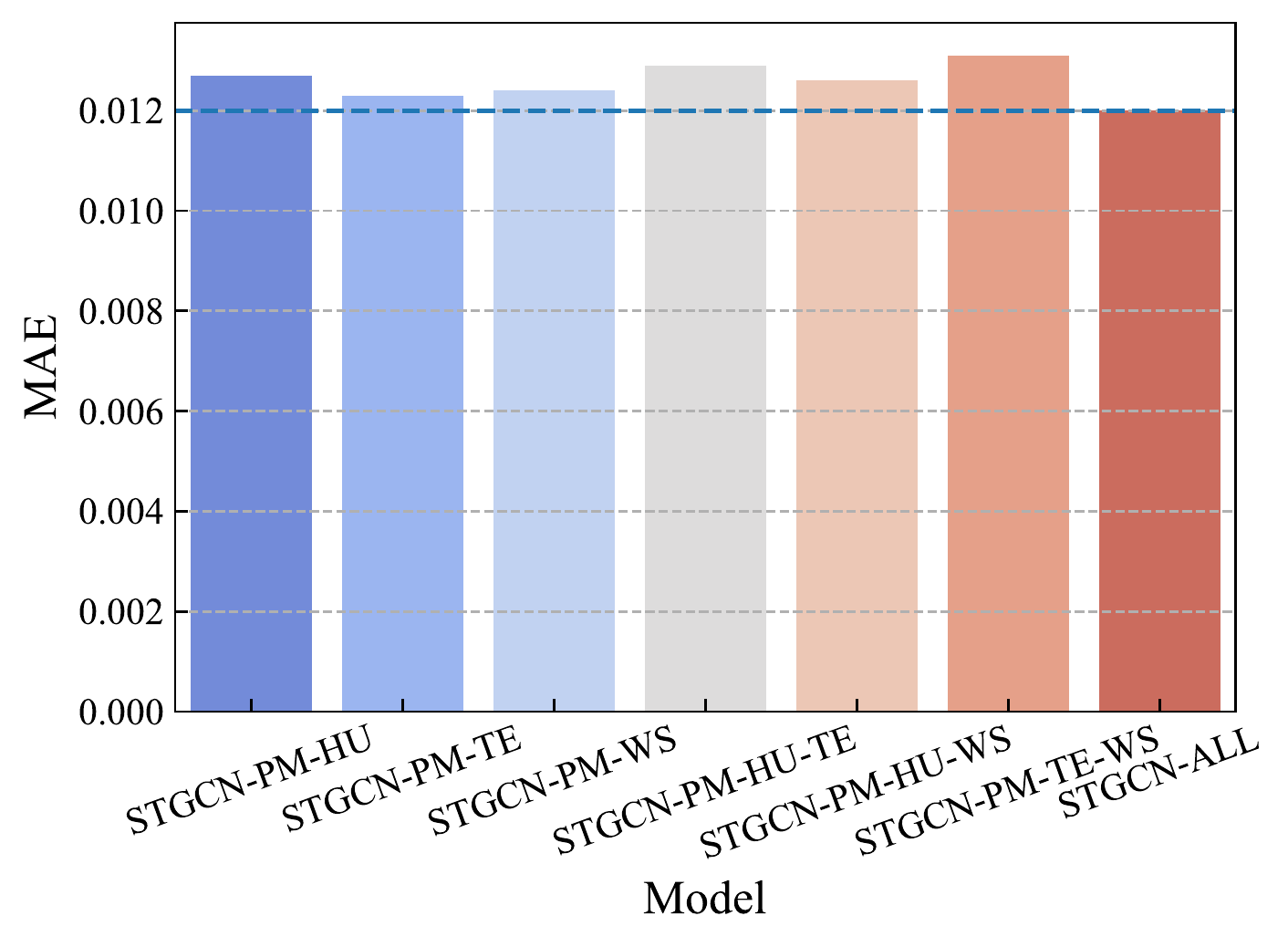}}
	\subfigure[RMSE]{
		\label{fig:subfig:16b} 
		\includegraphics[scale=0.5]{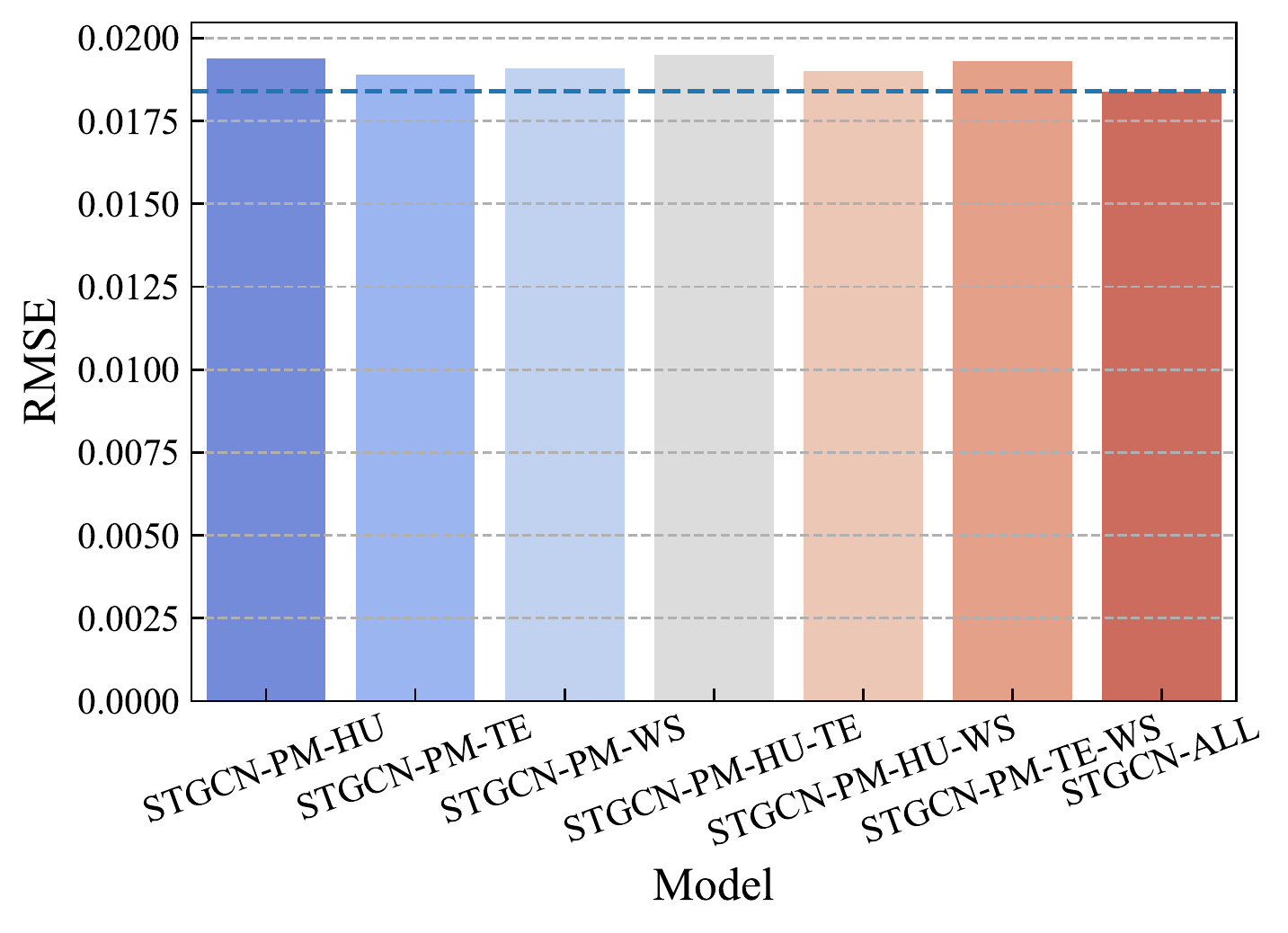}}
	\caption{Performance comparison of different combinations of observations} 
	\label{Figure16}
\end{figure}

Figure \ref{Figure16} demonstrates the different effects of different combinations of observations, including PM-HU (PM2.5 and humidity), PM-TE (PM2.5 and temperature), PM-WS (PM2.5 and wind speed), PM-HU-TE (PM2.5, humidity, and temperature), PM-HU-WS (PM2.5, humidity, and wind speed), PM-TE-WS (PM2.5, temperature, and wind speed), and ALL (all observations, i.e., K4). The results indicated that the predictions using all the fused information achieve the best performance.

\section{Discussion}
\label{sec:4:discussion}
In this paper, by considering the spatial correlations among monitoring points in specific application scenarios, we propose an RBF-based method to fuse heterogeneous data depending on the distances between monitoring points and present a novel deep learning method that uses a GCN to perform prediction when using processed data. In this specific air quality prediction scenario, monitoring points that belong to different monitoring sources and are scattered across different locations collect various observations over time. The results demonstrate the consistency and effectiveness of the fused data obtained by the proposed RBF-based method and further demonstrate the effectiveness of the employed deep learning method.

In this section, we discuss the advantages and shortcomings of the proposed method. Moreover, we discuss some potential future work that may be conducted to address these shortcomings.

\subsection{Advantages of the Proposed Method}

The proposed deep learning method achieves satisfactory prediction performance based on processed data (i.e., the data are first fused and then constructed; for details, see Section \ref{sec:2:methods}). The advantage and essential idea of the proposed method are the consideration of the spatial correlations in heterogeneous data collected from multiple monitoring points for specific application scenarios.

First, we consider the spatial distribution of the monitoring points scattered in a study area and fuse the heterogeneous data collected from the monitoring points with spatial information using the RBF-based data fusion method. This RBF-based fusion method has the following advantages. (1) Each monitoring point is able to obtain data depending on the distances between itself and its neighbors as well as the corresponding information from these neighbors. This implies that the RBF-based method allows for exploring multiple correlations among heterogeneous data according to their locations and expressing the complex interactions among external factors as spatial correlations. Then, the spatial correlations are incorporated into the distance-based fusion method, thus achieving highly accurate fusion. (2) The RBF-based method has a better interpretable result than those of other methods. (3) As a function dependent on distance, the RBF is suitable for the weighting and fusion of large amounts of data from multiple points scattered across a study area, thereby implying that the proposed model is highly applicable to heterogeneous data with explicit location information.

Second, we propose replacing the traditional single monitoring point or several monitoring points with graph-structured data. This approach has the following advantages. (1) The local information is replaced by global information. (2) When the fused information is embedded within the fully connected graph constructed from all the monitoring points, a prediction regarding the future trends for the entire study area can be achieved. (3) The necessity of selecting the appropriate monitoring points is eliminated, thus increasing the efficiency of the method. (4) The graph-structured data do not suffer from the situation where a limited number of monitoring points are adopted and information from other monitoring points is ignored.

Finally, we consider the capture of the spatial correlations between the constructed data by employing a state-of-the-art deep learning model (GCN). The GCN extracts the information stored at each node (i.e., monitoring point) through convolutional operations on the graph, and it is thus particularly applicable to non-grid data. Furthermore, we employ a deep learning model named STGCN that combines a GCNN (i.e., a deep learning model that can capture temporal correlations from the constructed sequence data) and the GCN (i.e., a deep learning model that can capture spatial correlations from the constructed graph-structured data) and thus achieves satisfactory prediction performance.

\subsection{Shortcomings of the Proposed Method}

There are three limitations of the proposed method. First, the method is designed specifically for the fusion of data from monitoring points with explicit location information. Therefore, it is not applicable to datasets without location information. Second, we only adopt the fully connected graph to construct the graph-structured data. In the above case, the real dataset consists of a limited of 53 nodes. Consequently, the interactions between nodes can be considered. Once a large number of nodes are distributed in a study area, the construction of a fully connected graph increases the computational cost of the method. Finally, regarding the deep learning predictions, the utilized dataset contains a limited 4 observations. The current results only indicate that the obtained predictions are best when employing all fused information. However, it is difficult to interpret the effects of the different combinations of only a few observations on the predictions of the model.

\subsection{Outlook and Future Work}

In the future, (1) we plan to evaluate the applicability of the proposed deep learning approach in other similar scenarios. Similar scenarios refer to the continuous collection of heterogeneous data from multiple monitoring stations scattered across a study area over time. The collected data include the relevant location information. “Applicability” refers to achieving satisfactory performance for spatiotemporal prediction utilizing fused data. (2) We also plan to employ the \textit{k}-Nearest Neighbors search algorithm (\textit{k}NN) to locally construct graph-structured data. This is because in some cases, there may be a large number of monitoring points, and the use of \textit{k}NN to construct the graph-structured data may significantly reduce the computational cost of the model. (3) We plan to employ other graph convolutional neural networks (e.g., GAT \cite{34}, GraphSAGE \cite{35}, Diffpool \cite{36}) to capture the spatial correlations among the fused data.

\section{Conclusion}
\label{sec:5:conclusion}
In this paper, we propose a deep learning method for fusing heterogeneous data collected from multiple monitoring points using a GCN to predict the future trends of some observations. The proposed method is applied in a real air quality prediction scenario. The essential idea behind the proposed method is to (1) fuse the heterogeneous data collected based on the locations of the monitoring points considering their spatial correlations and (2) perform prediction based on global information (i.e., fused information from all monitoring points in the entire study area) rather than local information (i.e., information from a single monitoring point or several monitoring points in a study area). In the proposed method, (1) a fusion matrix is assembled using RBF-based fusion; (2) a weighted adjacency matrix is constructed using Gaussian similarity functions; (3) a sequence of fused observation vectors is constructed based on the time series information of the above fusion matrix; and (4) the STGCN is employed to predict the future trends of several observations and is fed the above constructed data (i.e., the weighted adjacency matrix and the sequence of fused observation vectors). The results obtained on the real air quality dataset demonstrate that (1) the fused data derived from the RBF-based fusion method achieves satisfactory consistency; (2) the performances of the compared prediction models based on fused data are better than those based on raw data; and (3) the STGCN model achieves the best performance when compared with those of all baseline models. The proposed method is applicable for similar scenarios where continuous heterogeneous data are collected from multiple monitoring points scattered across a study area. Future work will focus on the application of the proposed method in scenarios where there are large numbers of monitoring points.

\section*{Acknowledgments}
This research was jointly supported by the National Natural Science Foundation of China (Grant Nos. 11602235), and the Fundamental Research Funds for China Central Universities (2652018091). The authors would like to thank the editor and the reviewers for their helpful comments and suggestions.

\section*{Reference}
  \bibliographystyle{elsarticle-num} 
  \bibliography{reference}
  

\end{document}